\documentclass[journal,twoside,web]{ieeecolor}
\usepackage{tmi}
\usepackage{cite}
\usepackage{amsmath,amssymb,amsfonts}
\usepackage{algorithmic}
\usepackage{graphicx}
\usepackage{textcomp}
\usepackage{booktabs}
\usepackage{multirow}
\usepackage{rotating}
\usepackage{array}       
\usepackage[ruled,linesnumbered]{algorithm2e}
\def\BibTeX{{\rm B\kern-.05em{\sc i\kern-.025em b}\kern-.08em
    T\kern-.1667em\lower.7ex\hbox{E}\kern-.125emX}}
\begin{document}

\title{Noise2Match: Self-Supervised Low-Dose CT Denoising via Adaptive Inter-Slice Matching}
\author{Langrui Zhou, \IEEEmembership{Graduate Student Member, IEEE}, Guang Li, \IEEEmembership{Member, IEEE}, Xinyu Huang, Ziteng Zhou, Zuyang Huang, Hongming Shan, \IEEEmembership{Senior Member, IEEE}, Shouhua Luo,  and Ning Gu
\thanks{
Manuscript received 2 May 2026; revised 21 July 2026 and 11 September 2026; accepted 7 October 2026. This work was supported in part by the National Key Research and Development Program of China under Grant 2022YFF0710800; in part by Jiangsu Provincial Frontier Technology Research and Development Program under Grant BF2025608; in part by Jiangsu Provincial Key Research and Development Program under Grant BE2021609; and in part by National Natural Science Foundation of China under Grant 62001112. Langrui Zhou and Guang Li contributed equally to this study. \textit{(Corresponding author: Guang Li.)}}
\thanks{Langrui Zhou, Guang Li, Xinyu Huang, Ziteng Zhou, Zuyang Huang, Shouhua Luo and Ning Gu are with the Jiangsu Key Laboratory for Biomaterials and Devices, School of Biological Science and Medical Engineering, Southeast University, Nanjing 210009, China. Ning Gu is also with the Jiangsu Key Laboratory for Cardiovascular Information and Health Engineering Medicine, Institute of Clinical Medicine, Nanjing Drum Tower Hospital, Medical School, Nanjing University, Nanjing 210093, China (e-mail: lrzhou@seu.edu.cn; liguang@seu.edu.cn; 230259211@seu.edu.cn; 220224428@aa.seu.edu.cn; zuyanghuang@seu.edu.cn; luoshouhua@seu.edu.cn; guning@seu.edu.cn).}
\thanks{Hongming Shan is with the Institute of Science and Technology for Brain-Inspired Intelligence, Fudan University, Shanghai 201203, China (e-mail: hmshan@fudan.edu.cn)}
}

\maketitle

\begin{abstract}
With the rapid development of deep learning, supervised convolutional neural network methods have greatly improved the performance of low-dose computed tomography (CT) denoising. However, they require paired noisy–clean training data, which is difficult to obtain in practice. Some classic self-supervised denoising methods rely on the assumption of pixel-wise independent noise, making them ineffective for correlated noise in low-dose CT (LDCT) images. Considering the high similarity and weaker noise correlation between adjacent CT slices, several recently proposed self-supervised methods use adjacent LDCT slices for training. Nevertheless, these approaches fail to effectively account for the non-rigid misalignment between slices caused by anatomical variation, hindering them from satisfying the equal-mean condition required by the classic Noise2Noise framework. In this paper, we propose Noise2Match, a self-supervised LDCT denoising method based on adaptive inter-slice matching. Noise2Match introduces an inter-slice matching network to model the label transition process between adjacent slices, and then applies correction to the denoiser’s outputs to better satisfy the equal-mean condition. We theoretically prove that, under idealized conditions, training Noise2Match using adjacent LDCT slices is equivalent to supervised training with paired noisy–clean images. We further find that Noise2Match can also effectively perform low-dose cone-beam CT (CBCT) denoising, which could be attributed to its correction operation that reduces inter-slice noise correlation. To the best of our knowledge, Noise2Match is the first effective image-domain self-supervised method for low-dose CBCT. Extensive experiments demonstrate that Noise2Match outperforms existing image-domain self-supervised methods, offering broader applicability and higher practical value.
\end{abstract}


\section{Introduction}
\label{sec_introduction}
\IEEEPARstart{C}{omputed} tomography (CT) plays a crucial role in clinical diagnosis. However, X-ray ionizing radiation during CT scanning exposes patients to increased health risks. Although reducing the tube current can lower the radiation dose, it introduces low-dose noise that degrades image quality and compromises clinical diagnosis. Therefore, denoising is of great importance in low-dose CT (LDCT) imaging. With the rapid development of deep learning, supervised denoising methods based on convolutional neural networks (CNNs) have greatly improved image denoising performance and have been widely applied to LDCT denoising \cite{SH19}. For example, Chen \emph{et al.} proposed a residual encoder–decoder convolutional neural network (RED-CNN) \cite{CH17} to learn the mapping from LDCT images to normal-dose CT (NDCT) images. Yang \emph{et al.} combined Wasserstein generative adversarial network (WGAN) with perceptual loss to develop WGAN-VGG \cite{YQ18}, which generates more realistic denoised images. These supervised denoising methods require aligned noisy-clean image pairs for training. However, in practical applications, acquiring such paired datasets is expensive, time-consuming, and sometimes infeasible \cite{LJ18}. Therefore, the practicality of CNN-based supervised LDCT denoising methods remains limited.

To eliminate the dependence on paired noisy–clean image data, the classic Noise2Noise framework \cite{LJ18} demonstrated that when the noise has zero mean, training a denoising network with another noisy observation of the same scene as the target yields performance comparable to that of using a clean target. However, in practical applications, it is often infeasible to acquire paired noisy–noisy images through repeated scans. Consequently, a series of self-supervised denoising methods that require only single noisy images for training have been proposed. Among them, blind-spot-based methods, such as Noise2Void \cite{KA19} and Blind2Unblind \cite{WZ22}, employ blind-spot structures that guide the model to predict the true value of a pixel from its neighboring pixels to achieve effective denoising. Data-splitting-based methods, such as Neighbor2Neighbor \cite{HT21} and Zero-shot Noise2Mean \cite{LD25}, create paired noisy sub-images by splitting a single noisy image, extending the Noise2Noise framework to scenarios with only single noisy images. However, these self-supervised methods typically assume pixel-wise independent noise, which limits their ability to suppress the correlated noise in LDCT images \cite{NC23}. To address this limitation, recent studies have focused on developing self-supervised approaches specifically designed for LDCT denoising \cite{ZY25}. Among them, filtering-guided methods, such as USGF \cite{WQ23}, AG-REDCNN \cite{HY25}, and Filter2Noise \cite{SY25}, design filtering mechanisms to extract prior information from LDCT images and guide the model to preserve structural details while suppressing noise. Frequency-separation-based methods, such as RIP-Net \cite{PJ24} and WIA-LD2ND \cite{ZH24}, decompose images into low-frequency components containing core structures and high-frequency components containing fine details and noise, and optimize them separately for self-supervised denoising. Noise-decoupling-based methods, such as SDCNN \cite{LY25}, decompose LDCT noise into signal-dependent and signal-independent components and handle them individually to improve overall denoising performance.

In recent years, several researchers have noticed that adjacent slices within the same CT volume exhibit distinctive properties \cite{JS23, NC23}. On one hand, anatomical structures are inherently continuous in space, and adjacent CT slices contain many corresponding regions belonging to the same tissues, which share equal underlying true values. On the other hand, due to the limited cross-slice interference during CT reconstruction, the noise is more independent between adjacent slices. This indicates that corresponding regions belonging to the same tissues in adjacent slices approximately satisfy the \textit{independent} and \textit{equal-mean} conditions required by Noise2Noise. Therefore, some researchers have been exploring how to leverage adjacent slices for effective self-supervised LDCT denoising. Jeon \emph{et al.} proposed MM-Net \cite{JS23}, which utilizes multiple adjacent LDCT slices to guide a multi-scale attention U-Net to ignore random noise within individual slices and focus on shared structural information across slices. Niu \emph{et al.} proposed a similarity-based self-supervised method, Noise2Sim \cite{NC23}, which computes the pixel-wise similarity between the current slice and one of its adjacent slices, excludes pixels with dissimilarity exceeding a predefined threshold, and optimizes the denoising model using only the remaining similar pixel pairs. Noise2Sim has been demonstrated to effectively remove correlated noise in LDCT images. Wang \emph{et al.} proposed a similarity-based visual blind-spot scheme (SVB) \cite{WY25}. SVB only uses adjacent slices with noise correlation below a set threshold as training data, and introduces blind-spot strategies into Noise2Sim to remove noise more adequately.

\begin{figure}[!t]
\centerline{\includegraphics[width=\columnwidth]{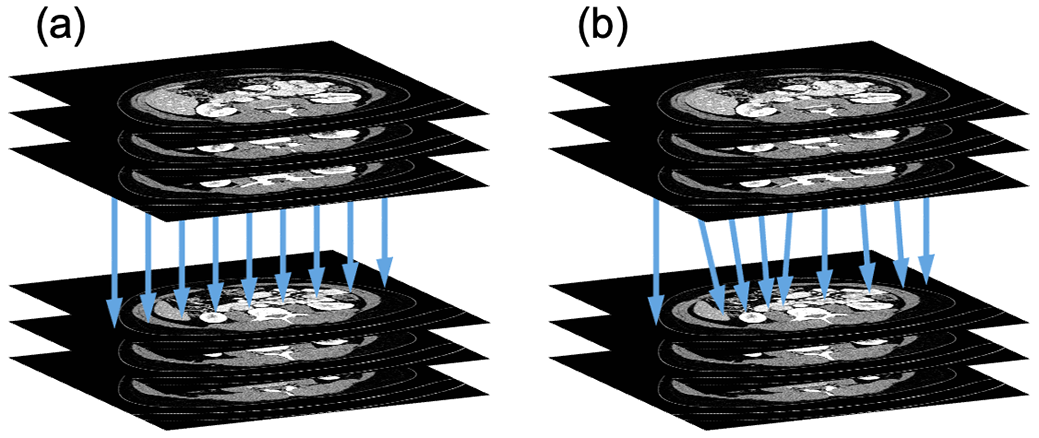}}
\caption{In self-supervised LDCT denoising based on adjacent slices, (a) directly computing the loss between same-coordinate pixels of adjacent slices does not conform to (b) the non-rigid inter-slice misalignment caused by anatomical variation.}
\label{fig1}
\end{figure}

Although these adjacent-slice-based self-supervised methods have demonstrated acceptable performance in LDCT denoising, they do not effectively satisfy the equal-mean condition required by Noise2Noise. For example, Noise2Sim assumes that same-coordinate regions of adjacent slices with dissimilarity below a set threshold belong to the same tissues, making it reasonable to compute pixel-wise loss within these regions (as shown in Fig.~\ref{fig1}(a)). However, due to longitudinal anatomical variation, the same tissue may exhibit non-rigid spatial misalignments between adjacent slices (as shown in Fig.~\ref{fig1}(b)). Therefore, even if the dissimilarity between same-coordinate regions is less than the threshold, there remains a high probability that they belong to different tissues. This indicates that directly computing loss between same-coordinate pixels cannot properly satisfy the equal-mean condition, and may lead to suboptimal constraints for the model. Similar issues also exist in other adjacent-slice-based self-supervised methods.

In fact, self-supervised training based on adjacent LDCT slices can be formulated as a problem of \textit{learning from unreliable (or noisy) labels}, which is an important topic in deep learning theory and aims to train robust models in the presence of annotation errors \cite{GE25, HK25}. Current approaches in this field mainly include: (1) \textit{label-transition-matrix-based methods} \cite{LZ25}, which model the label corruption or transition process and perform correction accordingly; (2) \textit{regularization-based methods} \cite{WH21}, which design specialized network architectures or training schemes to suppress fitting to incorrect labels; (3) \textit{loss-reweighting-based methods} \cite{ZL24}, which adjust the contribution of each sample to the loss function based on its estimated importance. Inspired by label-transition-matrix-based methods, we propose Noise2Match, a self-supervised LDCT denoising method based on adaptive inter-slice matching.  Noise2Match introduces an inter-slice matching network to model the label transition process between adjacent LDCT slices, and then applies the estimated transition matrix to correct the output of the denoiser, ensuring that pixels belonging to the same tissues are properly matched pairwise across adjacent slices. In this way, our method allows the transfer of the supervision signal between adjacent slices to adapt to non-rigid spatial misalignments caused by longitudinal anatomical variation. We theoretically prove that, under idealized conditions, training Noise2Match using adjacent LDCT slices is equivalent to supervised training with paired noisy–clean images. Experimental results demonstrate that Noise2Match outperforms both conventional denoising algorithms and state-of-the-art self-supervised deep learning methods in LDCT denoising. 

\begin{figure*}[!t]
\centerline{\includegraphics[width=\textwidth]{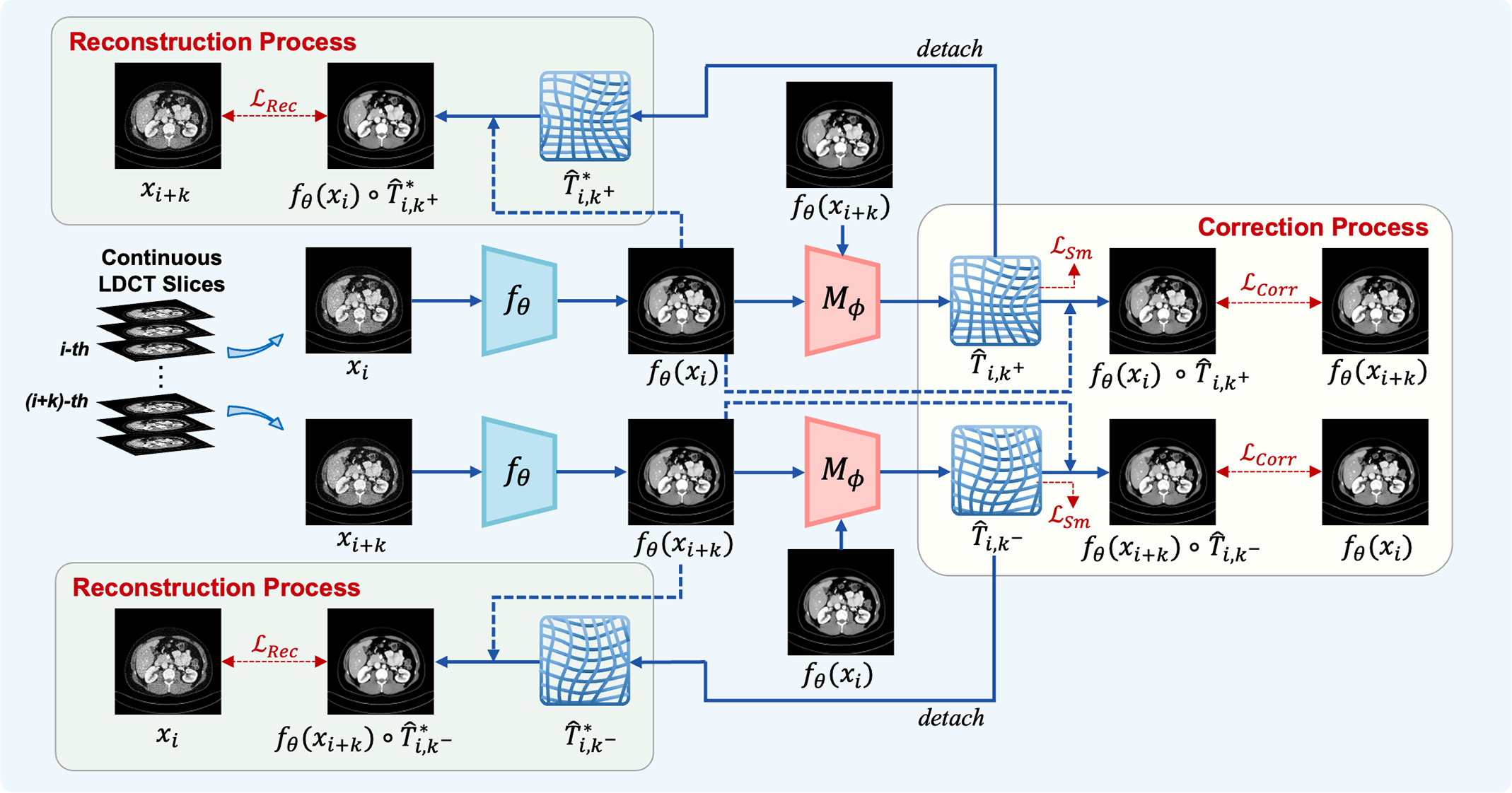}}
\caption{Overview of the proposed Noise2Match for LDCT denoising. The denoising network $f_\theta$ takes noisy LDCT images as input and predicts the corresponding clean images. The inter-slice matching network $M_\phi$ estimates the label transition matrix between adjacent slices based on the denoised results. The estimated transition matrix is then used to correct the denoiser's output.}
\label{fig2}
\end{figure*}

Furthermore, we find that Noise2Match can also effectively perform low-dose cone-beam CT (CBCT) denoising, which could be attributed to its correction operation that reduces inter-slice noise correlation. Considering that the optical crosstalk in flat-panel detector scintillators and 3D cone-beam reconstruction algorithms (e.g., FDK \cite{FDK84}) lead to higher inter-slice noise correlation in reconstructed CBCT images \cite{TD08, BJ11, ZH14}, most existing self-supervised CBCT denoising methods rely on projection-domain data \cite{CK23, LJ25}. Noise2Match provides a practical solution for performing self-supervised denoising directly in the image domain, making it highly promising for clinical applications.

In summary, the main contributions of this work are as follows. \textit{First}, we formulate self-supervised training based on adjacent LDCT slices as a problem of learning from unreliable labels and, based on this formulation, propose a novel adjacent-slice-based self-supervised LDCT denoising method, Noise2Match. \textit{Second}, we theoretically prove that training Noise2Match using adjacent LDCT slices is equivalent to supervised training using paired noisy–clean images under idealized conditions. \textit{Third}, experimental results on multiple datasets show that Noise2Match achieves superior performance compared to existing state-of-the-art self-supervised methods. \textit{More importantly}, to the best of our knowledge, Noise2Match is the first effective self-supervised method that can accomplish low-dose CBCT denoising solely in the image domain, which remains a challenging task for current image-domain self-supervised methods. Therefore, Noise2Match offers broader applicability and greater practical value. Our code is available at https://github.com/LangruiZhou/Noise2Match.


\section{Method}
\label{sec_method}
\subsection{Theoretical Motivation}
\subsubsection{Noise2Noise Revisit}
Noise2Noise \cite{LJ18} is a classic self-supervised denoising method that requires only paired independent noisy images of the same scene for model training. Specifically, given a pair of images $x_1=s+n_1$ and $x_2=s+n_2$ obtained from two independent noisy observations of the same scene, where $s$ denotes the underlying clean image, and $n_1$ and $n_2$ are independent zero-mean noise, Noise2Noise demonstrates that, in terms of expectation, training a denoising model $f_{\theta}$ to map a noisy image to another noisy image is equivalent to mapping it to the clean image, as expressed in Eq.~\eqref{eq1}:

\begin{small}
\begin{equation}
\arg\min_{\theta}\mathbb{E}_{x_1, x_2}[||f_{\theta}(x_1)-x_2||^2_2] = \arg\min_{\theta}\mathbb{E}_{x_1, s}[||f_{\theta}(x_1)-s||^2_2].
\label{eq1}
\end{equation}
\end{small}

\noindent Therefore, the model trained on paired noisy–noisy images in Noise2Noise can achieve denoising performance comparable to that of a supervised model trained on noisy–clean pairs. Adjacent-slice-based self-supervised LDCT denoising is essentially an implementation derived from the Noise2Noise framework \cite{JS23, NC23}, where the key idea is to construct approximately paired noisy training samples using adjacent slices. Adjacent LDCT slices exhibit a dual nature: on one hand, they are highly similar due to the spatial continuity of anatomical structures; on the other hand, longitudinal anatomical variation introduces non-rigid misalignments, making it challenging to fully satisfy the equal-mean condition required by Noise2Noise between same-coordinate pixels of adjacent slices. Consequently, the effectiveness of such methods critically depends on whether they can mitigate the interference in model optimization, which arises from mean differences between same-coordinate pixels of adjacent slices.

\subsubsection{Learning with Unreliable Labels}
If two adjacent slices with non-rigid misalignments are regarded as training labels containing annotation errors for each other, then self-supervised training based on adjacent slices can be formulated as a process of learning from unreliable labels. Specifically, given a LDCT volume $\mathbf{x}=\{x_i\}_{i=1}^N$ consisting of $N$ consecutive slices, where $x_i=s_i+n_i$ and $x_{i+k}=s_{i+k}+n_{i+k} (k\in Z^+)$ represent a pair of adjacent LDCT slices, $s_i$ and $s_{i+k}$ denote the corresponding underlying clean slices (i.e., ground truth), and $n_i$ and $n_{i+k}$ denote low-dose noise. Our goal is to train a denoiser $f_\theta$ using only paired data $\{x_i,x_{i+k}\}_{i=1}^{N-k}$ constructed from adjacent slices with non-rigid misalignments, while ensuring that its performance is comparable to that of a supervised model trained with aligned noisy-clean pairs $\{x_i,s_i\}_{i=1}^N$. As discussed earlier, directly optimizing the loss function defined in Eq.~\eqref{eq2} usually leads to compromised denoising performance of $f_\theta$, since the training process is affected by the misalignments between adjacent slices.

\begin{small}
\begin{equation}
\frac{1}{N-k}\sum_{i=1}^{N-k}||f_\theta(x_i)-x_{i+k}||_2^2.
\label{eq2}
\end{equation}
\end{small}

\noindent To address this issue, we reanalyzed the spatial characteristics of anatomical structures. We noticed that, due to the spatial continuity and gradual variation of tissues across adjacent slices, each pixel in the current slice generally has a corresponding counterpart in its adjacent slice. This mapping relationship can be described by a label transition matrix. Specifically, for any pair of adjacent slices $(x_i, x_{i+k})$, there exists a label transition matrix $T_{i,k^+}$ such that $s_{i+k}=s_i \circ T_{i,k^+}$. Here, $T_{i,k^+}$ essentially represents a deformation field, and “$\circ$” denotes the resampling operation applied to the image; for discrete images, the resampling induced by $T_{i,k^+}$ can be represented as a matrix operator, and $T_{i,k^+}$ is therefore referred to as a label transition matrix. If the label transition matrix $T_{i,k^+}$ between adjacent slices can be accurately estimated, the output of the denoiser $f_\theta$ can then be corrected accordingly, and the loss function of the model can be reformulated as Eq.~\eqref{eq3}.

\begin{small}
\begin{equation}
\frac{1}{N-k}\sum_{i=1}^{N-k}||f_\theta(x_i)\circ T_{i,k^+}-x_{i+k}||_2^2.
\label{eq3}
\end{equation}
\end{small}

\noindent Based on the corrected loss defined in Eq.~\eqref{eq3}, we derive the following Theorem 1, which establishes that, under idealized conditions, training the model on $\{x_i, x_{i+k}\}_{i=1}^{N-k}$ is equivalent to supervised training on the aligned noisy–clean dataset $\{x_i, s_i\}_{i=1}^{N}$.

\textbf{\textit{Theorem 1}}: Given a dataset of adjacent LDCT slice pairs $\{x_i,x_{i+k}\}_{i=1}^{N-k}$ and a dataset of paired noisy–clean images $\{x_i, s_i\}_{i=1}^{N}$, suppose that the noises $n_i$ and $n_{i+k}$ in $x_i$ and $x_{i+k}$ are independent and zero-mean, and that for pairs of locations in adjacent slices that can be matched through the label transition process, the corresponding noise values follow the same conditional distribution. If there exists a deformation field $T_{i,k^+}$ satisfying $s_{i+k}=s_i \circ T_{i,k^+}$, and the denoiser $f_\theta$ is equivariant with respect to $T_{i,k^+}$, i.e., $f_\theta(x_i)\circ T_{i,k^+}=f_\theta(x_i\circ T_{i,k^+})$, then, under the idealized setting in which images are modeled as two-dimensional continuous-domain signals, the following equivalence holds:

\begin{small}
\begin{equation}
\begin{split}
&\arg\min_{\theta}\mathbb{E}_{x_i, x_{i+k}}[||f_\theta(x_i)\circ T_{i,k^+}-x_{i+k}||_2^2]\\
=&\arg\min_{\theta}\mathbb{E}_{x_i,s_i}[||f_\theta(x_i)-s_i||_2^2].
\end{split}
\label{eq4}
\end{equation}
\end{small}

\noindent\textbf{\textit{Proof}}: According to the equivariance of $f_\theta$ with respect to $T_{i,k^+}$, we have

\begin{small}
\begin{equation}
\begin{split}
&\arg\min_{\theta}\mathbb{E}_{x_i, x_{i+k}}[||f_\theta(x_i)\circ T_{i,k^+}-x_{i+k}||_2^2]\\
=&\arg\min_{\theta}\mathbb{E}_{x_i,x_{i+k}}[||f_\theta(x_i\circ T_{i,k^+})-x_{i+k}||_2^2].
\end{split}
\label{eq5}
\end{equation}
\end{small}

\noindent Let $\tilde{x}_{i+k}=x_i\circ T_{i,k^+}=s_{i+k}+n_i\circ T_{i,k^+}$. The above objective can then be written as

\begin{small}
\begin{equation}
\arg\min_{\theta}\mathbb{E}_{\tilde{x}_{i+k},x_{i+k}}[||f_\theta(\tilde{x}_{i+k})-x_{i+k}||_2^2].
\label{eq6}
\end{equation}
\end{small}

\noindent Since $n_i$ and $n_{i+k}$ are independent and zero-mean, and the noise values at locations that can be spatially matched through the label transition process follow the same conditional distribution, the transformed noise $n_i\circ T_{i,k^+}$ and $n_{i+k}$ are independent and follow the same conditional distribution given $s_{i+k}$ under the idealized continuous-domain setting. Therefore, $\tilde{x}_{i+k}$ and $x_{i+k}$ can be regarded as two independent noisy observations of $s_{i+k}$, and they follow the same conditional distribution. By the Noise2Noise principle,

\begin{small}
\begin{equation}
\begin{split}
&\arg\min_{\theta}\mathbb{E}_{\tilde{x}_{i+k},x_{i+k}}[||f_\theta(\tilde{x}_{i+k})-x_{i+k}||_2^2]\\
=&\arg\min_{\theta}\mathbb{E}_{x_{i+k},s_{i+k}}[||f_\theta(x_{i+k})-s_{i+k}||_2^2]\\
=&\arg\min_{\theta}\mathbb{E}_{x_i,s_i}[||f_\theta(x_{i})-s_{i}||_2^2].
\end{split}
\label{eq7}
\end{equation}
\end{small}

It should be noted that the equivalence established in Theorem 1 relies on several idealized assumptions. In practical implementations, these assumptions can generally be satisfied only approximately because spatial transformations of discrete images involve interpolation and the transition matrix estimated by $M_\phi$ may contain errors. Therefore, in practice, the equivalence in Theorem 1 should be interpreted as approximate rather than strict.

\subsection{Noise2Match}
Based on the above theoretical motivation, we propose Noise2Match, a self-supervised LDCT denoising method based on adaptive inter-slice matching. Noise2Match consists of a denoising network $f_\theta$ and an inter-slice matching network $M_\phi$. To fully exploit information from adjacent slices and enhance the stability of the training process, we adopt a symmetric training scheme, as illustrated in Fig.~\ref{fig2}. The denoised outputs of two adjacent noisy slices $x_i$ and $x_{i+k}$ can be expressed as $f_\theta(x_i)$ and $f_\theta(x_{i+k})$.
The inter-slice matching network $M_\phi$ is responsible for estimating the label transition matrix between the latent clean slices $(s_i,s_{i+k})$ based on the denoised results $(f_\theta(x_i),f_\theta(x_{i+k}))$. 
The estimation result of the label transition matrix $T_{i,k^+}$ from $s_i$ to $s_{i+k}$ can be expressed as:

\begin{small}
\begin{equation}
\hat{T}_{i,k^+}=M_\phi(f_\theta(x_i), f_\theta(x_{i+k})).
\label{eq8}
\end{equation}
\end{small}

\noindent The estimation result of the label transition matrix $T_{i,k^-}$ from $s_{i+k}$ to $s_i$ can be expressed as:

\begin{small}
\begin{equation}
\hat{T}_{i,k^-}=M_\phi(f_\theta(x_{i+k}), f_\theta(x_i)).
\label{eq9}
\end{equation}
\end{small}

\noindent Here, both $\hat{T}_{i,k^+}$ and $\hat{T}_{i,k^-}$ have two channels representing the offsets in the x and y directions, respectively, and share the same width and height as the input images.

The denoiser $f_\theta$ and the matching network $M_\phi$ are optimized synchronously during the training process. This optimization consists of two mutually reinforcing parallel processes: reconstruction and correction. In the reconstruction process, the denoiser $f_\theta$ is optimized using the corrected data to enhance its denoising capability, thereby providing cleaner reference images for the correction process. In the correction process, the inter-slice matching network $M_\phi$ is optimized based on the denoising results of $f_\theta$ to estimate more accurate label transition matrices, which in turn further improve the reconstruction process. It should be noted that the inter-slice matching network $M_\phi$ and the correction operation are used only during training. During inference, only the denoising network $f_\theta$ is retained, and no spatial warping or resampling is applied to the denoised output.

\subsubsection{Reconstruction}
The reconstruction process is derived from the classic Noise2Noise framework. To better satisfy the equal-mean condition required by Noise2Noise, we correct the denoising outputs $f_\theta(x_i)$ and $f_\theta(x_{i+k})$ using the label transition matrices predicted by $M_\phi$, according to Theorem 1. The reconstruction loss can then be expressed as Eq.~\eqref{eq10}.

\begin{small}
\begin{equation}
\frac{1}{N-k}\sum_{i=1}^{N-k}[||f_\theta(x_i)\circ \hat{T}_{i,k^+}^{*}-x_{i+k}||_2^2+||f_\theta(x_{i+k})\circ \hat{T}_{i,k^-}^*-x_i||_2^2].
\label{eq10}
\end{equation}
\end{small}

\noindent Here, $\hat{T}_{i,k^+}^{*}$ and $\hat{T}_{i,k^-}^{*}$ denote detached copies of $\hat{T}_{i,k^+}$ and $\hat{T}_{i,k^-}$, respectively, without gradient information. We do not use the original $\hat{T}_{i,k^+}$ and $\hat{T}_{i,k^-}$ with gradients because there are two potential optimization paths to minimize the reconstruction loss: one is to optimize $f_\theta$ for improved denoising performance, and the other is to optimize $M_\phi$ to predict better label transition matrices to fit the reconstruction loss. To avoid conflicts between these two paths when minimizing the reconstruction loss, we use the detached $\hat{T}_{i,k^+}^{*}$ and $\hat{T}_{i,k^-}^{*}$ in Eq.~\eqref{eq10}. It is worth noting that during the early stages of training, when the model performance is still limited, $M_\phi$ may not be able to accurately estimate the label transition matrices between adjacent slices. However, this does not hinder the optimization of $f_\theta$. Benefiting from the high similarity between adjacent slices, $f_\theta$ can still learn approximate noise suppression patterns directly from adjacent slices based on the reconstruction loss, allowing it to continuously improve its denoising performance.

\subsubsection{Correction}
In the correction process, we adopt the correction loss as defined in Eq.~\eqref{eq11}.

\begin{small}
\begin{equation}
\begin{split}
\frac{1}{N-k}\sum_{i=1}^{N-k}[&||f_\theta(x_i)\circ \hat{T}_{i,k^+}-f_\theta(x_{i+k})||_2^2\\
+&||f_\theta(x_{i+k})\circ \hat{T}_{i,k^-}-f_\theta(x_i)||_2^2].
\label{eq11}
\end{split}
\end{equation}
\end{small}

\noindent The correction loss serves as a shared constraint for both $f_\theta$ and $M_\phi$, aiming to guide the model in seeking the optimal solution for both the image denoising task and the inter-slice matching task. For the denoiser $f_\theta$, the correction loss can be regarded as a regularization constraint on the reconstruction loss (Eq.~\eqref{eq10}). It enhances the stability of $f_\theta$ by requiring consistent denoising results for regions belonging to the same tissues. For the inter-slice matching network $M_\phi$, by correcting the denoising result of the current slice to match its adjacent slice’s counterpart, the correction loss guides $M_\phi$ to adaptively fit the label transition process. Moreover, since $x_{i+k}$ and $x_i\circ T_{i,k^+}$ correspond to the same underlying clean slice $s_{i+k}$, where $T_{i,k^+}$ denotes the naturally existing label transition matrix, $f_\theta(x_{i+k})$ can be regarded as the corresponding denoising result of $x_i\circ T_{i,k^+}$. Therefore, by enforcing consistency between $f_\theta(x_i)\circ \hat{T}_{i,k^+}$ and $f_\theta(x_{i+k})$, the correction loss encourages the denoiser $f_\theta$ to approximately satisfy the equivariance with respect to $T_{i,k^+}$ required by Theorem 1. In addition, to prevent the deformation fields output by $M_\phi$ from being excessively distorted or unnatural, following the approach of \cite{BG19}, we introduce a smooth loss $\mathcal{L}_{Sm}$ constructed using an $L_2$ penalty on the spatial gradients to regularize $\hat{T}_{i,k^+}$ and $\hat{T}_{i,k^-}$, as shown in Eq.~\eqref{eq12}.

\begin{small}
\begin{equation}
\mathcal{L}_{Sm}=\frac{1}{N-k}\sum_{i=1}^{N-k}[||\nabla\hat{T}_{i,k^+}||^2+||\nabla\hat{T}_{i,k^-}||^2].
\label{eq12}
\end{equation}
\end{small}

\subsubsection{Loss Reweighting for Mismatched Pixels}
When a slice lies at the longitudinal boundary of certain anatomical structures, a small number of pixels in the adjacent slice may be mismatched; that is, a few pixels in the current slice cannot find corresponding pixels with the same true value within the neighborhood of the adjacent slice. To mitigate the interference of these mismatched pixels on model optimization, we further apply reweighting to both the reconstruction loss and the correction loss. Specifically, when computing the reconstruction loss, we first use a fixed threshold $th$ to identify regions where significant differences remain between the corrected denoising results of adjacent slices, and mark these regions as zero in the weight matrices $w_{i,k^+}$ and $w_{i,k^-}$, i.e.,

\begin{equation}
w_{i,k^+}^{(u,v)}=
\begin{cases}
0, &|f_\theta(x_i)\circ \hat{T}_{i,k^+}-f_\theta(x_{i+k})|^{(u,v)}>th\\
1, &otherwise
\end{cases}
,
\label{eq13}
\end{equation}

\begin{equation}
w_{i,k^-}^{(u,v)}=
\begin{cases}
0, &|f_\theta(x_{i+k})\circ \hat{T}_{i,k^-}-f_\theta(x_i)|^{(u,v)}>th\\
1, &otherwise
\end{cases}.
\label{eq14}
\end{equation}

\noindent Here, $(u,v)$ denotes any position in the image. Subsequently, the reconstruction loss in Eq.~\eqref{eq10} is multiplied pixel-wise by $w_{i,k^+}$ and $w_{i,k^-}$ to obtain the final reconstruction loss $\mathcal{L}_{Rec}$:

\begin{equation}
\begin{split}
\mathcal{L}_{Rec}=\frac{1}{N-k}\sum_{i=1}^{N-k}[&||(f_\theta(x_i)\circ \hat{T}_{i,k^+}^*-x_{i+k})\odot w_{i,k^+}||_2^2 \\
+&||(f_\theta(x_{i+k})\circ \hat{T}_{i,k^-}^*-x_i)\odot w_{i,k^-}||_2^2].
\end{split}
\label{eq15}
\end{equation}

\noindent For the correction loss, considering that pixel mismatches between adjacent slices may also arise from prediction errors of the matching network $M_\phi$, we did not adopt the same zero-masking strategy as in the reconstruction loss. Instead, we constrained the correction loss values within the threshold $th$, allowing $M_\phi$ to continue optimization in a mild manner. The final correction loss $\mathcal{L}_{Corr}$ is formulated as follows: 

\begin{small}
\begin{equation}
\begin{split}
\mathcal{L}_{Corr}=\frac{1}{N-k}\sum_{i=1}^{N-k}[&||\min(|f_\theta(x_i)\circ \hat{T}_{i,k^+}-f_\theta(x_{i+k})|,th)||_2^2\\
+ &||\min(|f_\theta(x_{i+k})\circ \hat{T}_{i,k^-}-f_\theta(x_i)|,th)||_2^2].
\end{split}
\label{eq16}
\end{equation}
\end{small}

Finally, the overall optimization objective can be formulated as Eq.~\eqref{eq17}, where $\lambda_{Corr}$ and $\lambda_{Sm}$ are hyperparameters controlling the strengths of $\mathcal{L}_{Corr}$ and $\mathcal{L}_{Sm}$, respectively. The complete training procedure is shown in Alg.~\ref{alg1}.

\begin{small}
\begin{equation}
\mathcal{L}=\mathcal{L}_{Rec}+\lambda_{Corr}\cdot\mathcal{L}_{Corr}+\lambda_{Sm}\cdot\mathcal{L}_{Sm}.
\label{eq17}
\end{equation}
\end{small}

\begin{small}
\begin{algorithm}[!t]
\caption{Training for Noise2Match}
\label{alg1}
\KwIn{Dataset composed of adjacent LDCT slices $X=\{(x_i,x_{i+k})\}_{i=1}^{N-k}$; denoising network $f_\theta$; inter-slice matching network $M_\phi$.}
\While{not converged}{
\textbf{Sample} a pair of adjacent LDCT images $(x_i, x_{i+k})\in X$\;
\textbf{Derive} denoised images $f_\theta(x_i)$ and $f_\theta(x_{i+k})$\;
\textbf{Derive} estimation of label transition matrices:
$\hat{T}_{i,k^+}\leftarrow M_\phi(f_\theta(x_i),f_\theta(x_{i+k}))$,\
$\hat{T}_{i,k^-}\leftarrow M_\phi(f_\theta(x_{i+k}), f_\theta(x_i))$\; 

\textbf{Calculate} $\mathcal{L}_{Corr}$ by Eq.~\eqref{eq16} and $\mathcal{L}_{Sm}$ by Eq.~\eqref{eq12}\;
\textbf{Detach} $\hat{T}_{i,k^+}$ and $\hat{T}_{i,k^-}$ from computation graph:
$\hat{T}_{i,k^+}^{*}\leftarrow detach(\hat{T}_{i,k^+}), \hat{T}_{i,k^-}^{*}\leftarrow detach(\hat{T}_{i,k^-})$\;
\textbf{Derive} weight matrices $w_{i,k^+}$ and $w_{i,k^-}$ by Eq.~\eqref{eq13} and \eqref{eq14}, respectively\;
\textbf{Calculate} $\mathcal{L}_{Rec}$ by Eq.~\eqref{eq15}\;
\textbf{Update} $f_\theta$ and $M_\phi$ by minimizing the objective $\mathcal{L}=\mathcal{L}_{Rec}+\lambda_{Corr}\cdot\mathcal{L}_{Corr}+\lambda_{Sm}\cdot\mathcal{L}_{Sm}$
}
\end{algorithm}
\end{small}


\section{Experiments and Results}
\label{sec_exp}
\subsection{Datasets}
\subsubsection{Mayo 2016 Dataset}
We utilized the “2016 NIH-AAPM-Mayo Clinic Low-Dose CT Grand Challenge” dataset \cite{MC17} for training and testing, which contains 5,936 pairs of noisy-clean CT images from 10 patients with a slice thickness of 1 mm. Unless otherwise specified, we randomly selected data from 9 patients for training and used the remaining 1 patient for testing. The training set included 5,326 slices, while the test set contained 610 slices. The image size in this dataset is 512 $\times$ 512.

\subsubsection{Mayo 2020 Dataset}
To further validate the performance of different methods on clinical low-dose CT datasets, we additionally utilized the “Low-Dose CT Image and Projection Dataset” released by the Mayo Clinic in 2020 \cite{Mayo2020}. This dataset contains paired noisy-clean CT scans of the head, abdomen, and chest from 299 patients. We selected the abdominal CT data for training and testing. The training set consisted of 7,293 slices from 45 patients, while the test set included 866 slices from 5 patients. The image size in this dataset is 512 $\times$ 512.

\subsubsection{Piglet Dataset}
To evaluate the denoising performance of different methods across various types of datasets, we further used the piglet dataset \cite{YX18} acquired with a GE Discovery CT 750 HD scanner, which contains 800 CT slices from a piglet, each with a pixel size of 0.63$\times$0.63 mm$^2$ and a thickness of 0.625 mm. For each slice, the dataset provides scans at normal dose, 50\% dose, 25\% dose, 10\% dose, and 5\% dose levels. In our experiments, we used the 25\% dose images to train self-supervised models, and the normal-dose images were used to assess their denoising performance. The image size in this dataset is 512 $\times$ 512.

\subsubsection{Real-World CBCT Dataset}
To evaluate the performance of different methods on the low-dose CBCT denoising task, we performed scans on two rats using a Hiscan XM cone-beam micro-CT scanner, with an isotropic resolution of 13 $\mu$m, a tube voltage of 60 kV, a tube current of 134 $\mu$A, and an angular interval of 0.5 degrees. The training set consisted of 1,167 low-dose CBCT images from one rat, while the test set included 1,167 low-dose CBCT images from the other. Corresponding NDCT images were not provided in this dataset. Each image was cropped to a size of 1152 $\times$ 1152.

\subsubsection{ICASSP CBCT Dataset}
To further validate the performance of various methods on clinical low-dose CBCT datasets, we additionally utilized the dataset from the “2024 ICASSP Low-Dose CBCT Reconstruction Challenge” \cite{ICASSP}. Utilizing the LIDC-IDRI chest CT dataset \cite{LIDC-IDRI} as the ground truth, this dataset simulates clinical low-dose CBCT data for 1,010 patients (the number of incident photons is set to $10^5$). We selected 5,120 slices from 20 patients to construct the training set, and selected 768 slices from 3 patients to construct the test set. The image size in this dataset is 256 $\times$ 256.

\begin{figure*}[!t]
\centerline{\includegraphics[width=0.9\textwidth]{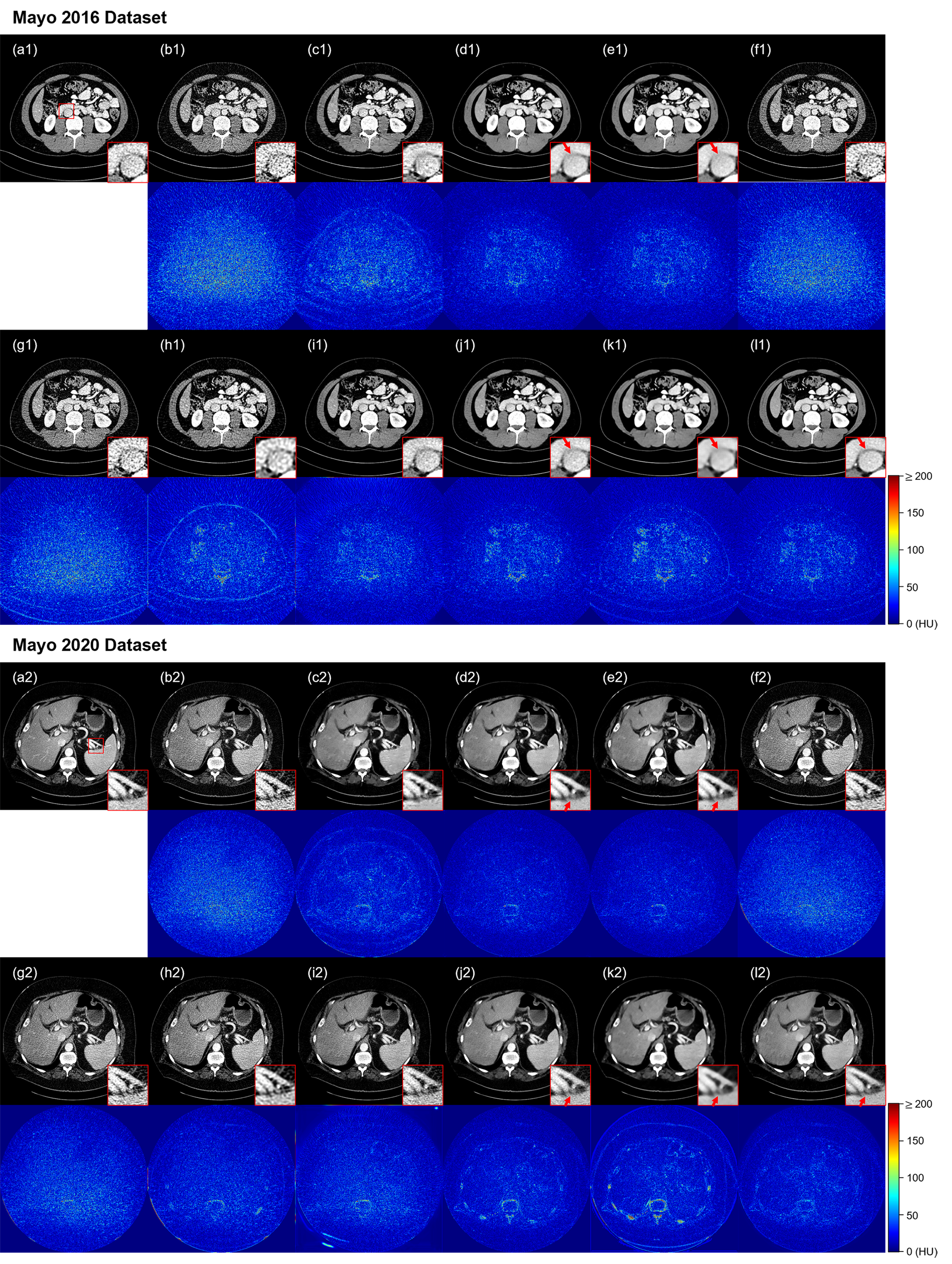}}
\caption{Qualitative results of different methods on the Mayo 2016 and Mayo 2020 datasets. (a) NDCT (ground truth), (b) LDCT, (c) BM3D, (d) RED-CNN, (e) Noise2Clean, (f) Noise2Void, (g) Neighbor2Neighbor, (h) SDCNN, (i) MM-Net, (j) Noise2Sim, (k) SVB, (l) Noise2Match (\textbf{ours}). }
\label{fig3}
\end{figure*}

\begin{table*}[!t]
\centering
\caption{Quantitative results (mean$\pm$std) of different methods on the Mayo 2016 and Mayo 2020 datasets.}
\resizebox{0.8\textwidth}{!}{
\begin{tabular}{lcccccc}
     \toprule
     \multirow{2}{*}{Methods} & \multicolumn{3}{c}{Mayo 2016} & \multicolumn{3}{c}{Mayo 2020} \\
     \cmidrule(r){2-4} \cmidrule(l){5-7}
     & PSNR (dB)$\uparrow$ & SSIM$\uparrow$ & RMSE (HU)$\downarrow$ & PSNR (dB)$\uparrow$ & SSIM$\uparrow$ & RMSE (HU)$\downarrow$ \\
     \midrule
     LDCT & 34.75$\pm$0.95 & 0.8211$\pm$0.0342 & 36.81$\pm$4.23 & 36.44$\pm$3.04 & 0.8682$\pm$0.0662 & 26.68$\pm$11.29 \\
     BM3D & 37.25$\pm$0.88 & 0.8952$\pm$0.0253 & 27.58$\pm$2.91 & 39.03$\pm$2.80 & 0.8977$\pm$0.0291 & 25.21$\pm$6.00 \\
     \midrule
     RED-CNN & 40.56$\pm$0.89 & 0.9435$\pm$0.0123 & 18.83$\pm$1.99 & 43.54$\pm$2.85 & 0.9691$\pm$0.0184 & 14.16$\pm$5.95 \\
     Noise2Clean & 40.52$\pm$0.90 & 0.9437$\pm$0.0127 & 18.93$\pm$2.01 & 43.41$\pm$2.95 & 0.9698$\pm$0.0173 & 14.44$\pm$6.49\\
     \midrule
     Noise2Void & 35.42$\pm$0.91 & 0.8420$\pm$0.0317 & 34.06$\pm$3.76 & 38.35$\pm$2.97 & 0.8762$\pm$0.0582 & 26.13$\pm$10.96\\
     Neighbor2Neighbor & 35.86$\pm$0.89 & 0.8541$\pm$0.0302 & 32.37$\pm$3.47 & 38.87$\pm$3.35 & 0.8850$\pm$0.0608 & 25.81$\pm$12.22\\
     SDCNN & 37.63$\pm$0.77 & 0.8934$\pm$0.0236 & 26.35$\pm$2.43 & 39.43$\pm$2.99 & 0.9356$\pm$0.0407 & 22.84$\pm$9.82\\
     MM-Net & 38.29$\pm$0.85 & 0.9112$\pm$0.0255 & 24.45$\pm$2.51 & 39.63$\pm$2.37 & 0.9250$\pm$0.0396 & 23.33$\pm$8.08\\
     Noise2Sim & 38.75$\pm$0.68 & 0.9367$\pm$0.0161 & 23.14$\pm$1.80 & 40.45$\pm$2.56 & 0.9574$\pm$0.0314 & 20.18$\pm$7.25\\
     SVB & 39.01$\pm$0.78 & 0.9369$\pm$0.0173 & 22.48$\pm$2.01 & 40.74$\pm$2.10 & 0.9576$\pm$0.0168 & 19.29$\pm$6.33 \\
     Noise2Match (Ours) & 40.08$\pm$0.84 & 0.9410$\pm$0.0167 & 19.89$\pm$1.96 & 42.88$\pm$2.48 & 0.9648$\pm$0.0189 & 15.32$\pm$6.46 \\
     \bottomrule
\end{tabular}
}
\label{tab1}
\end{table*}

\subsection{Implementation Details}
The denoising network $f_\theta$ was implemented based on U-Net \cite{RO15} with a residual connection. Following \cite{ZL24}, the inter-slice matching network $M_\phi$ was implemented as a U-Net variant with both input and output channel numbers set to 2. The model was optimized using the Adam optimizer, with a learning rate of $5\times10^{-4}$. The number of training epochs was 80, and the batch size was set to 4. Noise2Match was implemented in PyTorch and trained on one NVIDIA GeForce RTX 4090 GPU (24 GB). The hyperparameters in the loss function were set to $\lambda_{Corr}=1$ and $\lambda_{Sm}=0.001$. The threshold $th$ used for loss reweighting was set to 80 HU. The parameter $k$ was set to 1 for the Mayo 2016, Mayo 2020, Piglet, and ICASSP CBCT datasets, and to 2 for the Real-World CBCT dataset. The strategy for selecting $k$ is described in detail in Section~\ref{sec_k}. All images were normalized to the range $[0,1]$ before being fed into the network.

We compared the proposed Noise2Match with three different types of denoising methods: (1) \textit{traditional method}: Three-dimensional Block Matching (BM3D) \cite{DK07}; (2) \textit{supervised methods}: RED-CNN \cite{CH17} and Noise2Clean \cite{RO15}; (3) \textit{self-supervised methods}: Noise2Void \cite{KA19}, Neighbor2Neighbor \cite{HT21}, SDCNN \cite{LY25}, MM-Net \cite{JS23}, Noise2Sim \cite{NC23} and SVB \cite{WY25}. Among these, Noise2Clean employed the same network architecture as the denoising network of Noise2Match and was trained in a supervised manner using paired noisy–clean data. The remaining comparative methods were implemented following the settings provided in their original papers or official open-source codes. To quantitatively evaluate the denoising performance of the different methods, we employed three commonly used image quality metrics: peak signal-to-noise ratio (PSNR), structural similarity (SSIM), and root mean squared error (RMSE). All quantitative denoising performance metrics reported for Noise2Match were computed directly from the denoiser output $f_\theta$, without using the inter-slice matching network $M_\phi$, correction operation, or any spatial resampling during inference.

\subsection{Performance Comparison on Clinical LDCT Datasets}
In this section, we trained different deep learning models on the training sets of the Mayo 2016 and Mayo 2020 datasets, and evaluated the performance of all methods on their corresponding test sets. Fig.~\ref{fig3} presents representative image results from different methods, along with their residual maps relative to the NDCT images (ground truth). BM3D exhibited poor content fidelity, with many erroneous structures in the resulting images. The supervised methods, RED-CNN and Noise2Clean, effectively suppressed low-dose noise but tended to over-smooth the images, leading to the loss of structural details, as evidenced by the blurred tissue structures marked by red arrows in the regions of interest (ROIs) in Fig.~\ref{fig3}(d) and (e). From the residual maps, it is evident that Noise2Void and Neighbor2Neighbor barely suppressed the low-dose noise, which can be attributed to the correlated low-dose noise violating the assumption of pixel-wise independent noise. SDCNN achieved a certain degree of noise suppression through its noise-decoupling mechanism. However, the noise removal was still insufficient, with residual granular textures still visible in the denoised images, which also appeared slightly smoothed. Among the adjacent-slice-based self-supervised methods, MM-Net effectively mitigated low-dose noise, but some residual noise persisted in its resulting images, and slight CT value shifts could be observed in the background areas of the residual maps. Compared to MM-Net, Noise2Sim and SVB exhibited superior denoising capabilities. However, minor residual noise was still observable in the results of Noise2Sim, and the outputs of SVB were slightly over-smoothed. As indicated by the red arrows in the ROIs in Fig.~\ref{fig3}(j) and (k), both Noise2Sim and SVB had limitations in preserving fine tissue structures. Among all self-supervised methods, only Noise2Match demonstrated clear structural details within the ROIs, which may be attributed to its better adherence to the equal-mean condition. The above experimental results indicate that the proposed Noise2Match not only effectively removes low-dose noise but also excels in preserving textures and maintaining structural details.

Table~\ref{tab1} presents the quantitative results of all methods on the test sets. Our Noise2Match outperformed all traditional and self-supervised methods. Among the self-supervised methods, Noise2Match achieved the performance closest to that of the supervised methods. To further assess the statistical significance of the improvements over the strongest self-supervised baselines, we performed two-sided paired Wilcoxon signed-rank tests between Noise2Match and Noise2Sim/SVB using the per-slice PSNR, SSIM, and RMSE values, with statistical significance defined as $p<0.05$. On both the Mayo 2016 and Mayo 2020 datasets, the improvements of Noise2Match over Noise2Sim and SVB were statistically significant for all three metrics (all $p<0.001$). These results thoroughly demonstrate the exceptional capability of the proposed Noise2Match method in removing low-dose noise from clinical LDCT images.

\begin{figure}[!t]
\centerline{\includegraphics[width=\columnwidth]{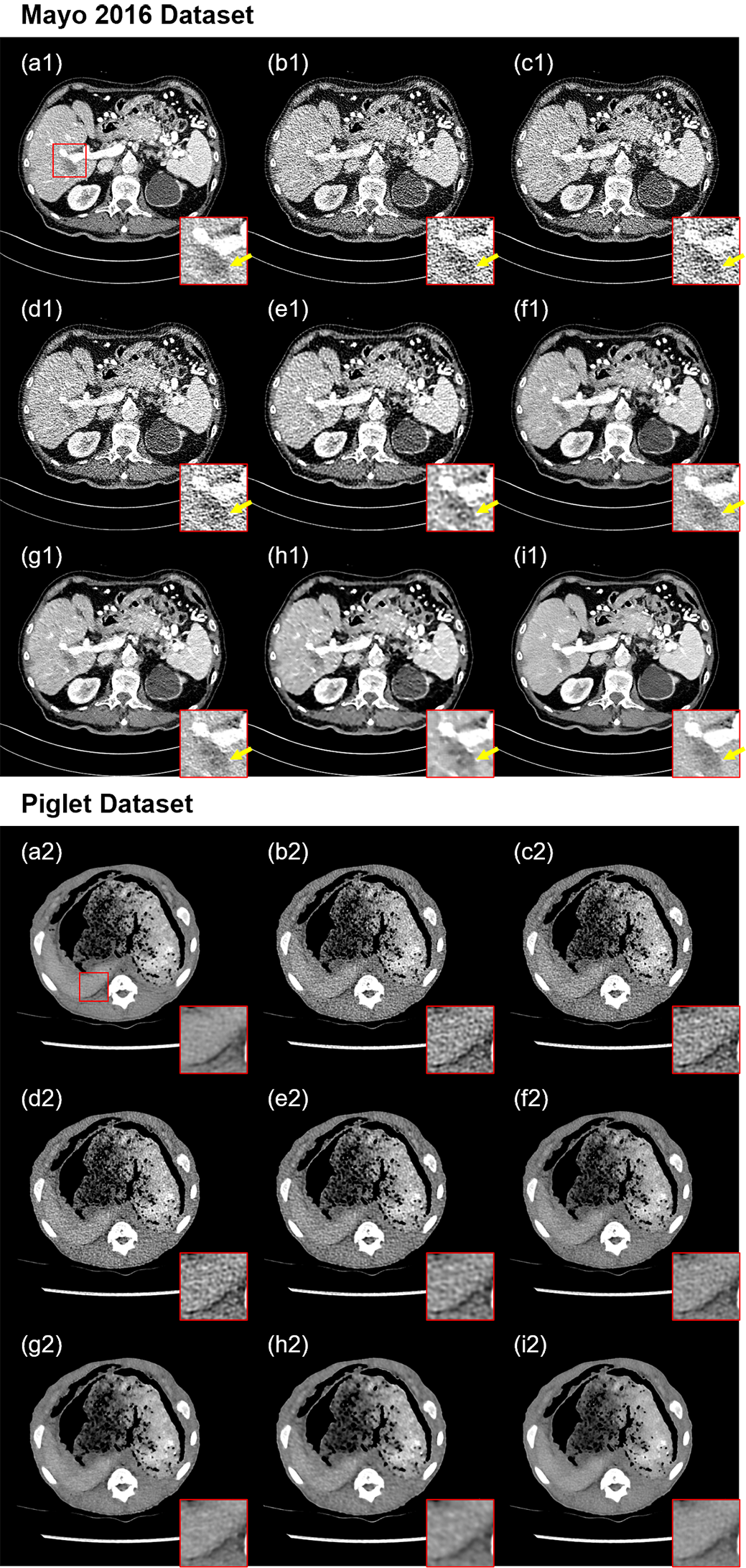}}
\caption{Qualitative results of different self-supervised methods on the Mayo 2016 and Piglet datasets. These models are all directly trained on the corresponding single-volume LDCT data. (a) NDCT (ground truth), (b) LDCT, (c) Noise2Void, (d) Neighbor2Neighbor, (e) SDCNN, (f) MM-Net, (g) Noise2Sim, (h) SVB, (i) Noise2Match (\textbf{ours}). }
\label{fig4}
\end{figure}

\subsection{Performance Comparison on Single LDCT Volume}
In some special scenarios, particularly when rare lesions with distributions substantially different from those of normal tissues are present, denoising models trained on CT data from other patients often exhibit limited generalization to these specific lesions or anatomical structures. Meanwhile, the scarcity of such special samples makes it difficult to obtain sufficient dedicated training data. Therefore, performing denoising directly on a single LDCT volume using self-supervised methods holds important research significance and practical value. In this section, we additionally selected data from a patient with liver lesions in the Mayo 2016 Dataset and further used the Piglet Dataset. Both datasets contain paired LDCT and NDCT images from a single CT volume. We first trained different self-supervised models using the LDCT images from each dataset and then evaluated their denoising performance based on the corresponding NDCT images. Fig.~\ref{fig4} presents representative results of different methods on both datasets. It can be observed that Noise2Void and Neighbor2Neighbor still failed to effectively suppress the correlated noise in LDCT images. SDCNN was able to suppress the noise to some extent, but its overall performance remained inferior to that of the adjacent-slice-based self-supervised baselines, and its ability to preserve key structures, such as low-contrast lesions, was relatively limited. MM-Net generally preserved the overall anatomical structures well, but noticeable residual noise remained in its outputs. Noise2Sim and SVB achieved stronger noise suppression; however, as indicated by the yellow arrows in the ROIs, the lesion boundaries in their denoised results remained relatively indistinct. In contrast, our proposed Noise2Match not only effectively suppressed low-dose noise but also better preserved image content and structural information, while providing improved boundary preservation for the low-contrast liver lesions in the Mayo 2016 Dataset.

The quantitative results of different self-supervised methods on the two datasets are listed in Table~\ref{tab2}. Noise2Match achieved superior quantitative performance compared with the other self-supervised methods on both the Mayo 2016 Dataset and the Piglet Dataset. These qualitative and quantitative results jointly demonstrate that Noise2Match has clear advantages over existing self-supervised methods for single-volume LDCT denoising.

\begin{table}
\centering
\caption{Quantitative results (mean$\pm$std) of different self-supervised methods on the Mayo 2016 and Piglet datasets. These models are all directly trained on the corresponding single-volume LDCT data.}
\resizebox{\columnwidth}{!}{
\begin{tabular}{clccc}
    \toprule
     & Methods & PSNR (dB)$\uparrow$ & SSIM$\uparrow$ & RMSE (HU)$\downarrow$  \\
    \midrule
    \multirow{6}{*}{\rotatebox{90}{Mayo 2016}} &LDCT & 36.39$\pm$1.86 & 0.8605$\pm$0.0493 & 31.06$\pm$7.62\\
    & Noise2Void & 36.81$\pm$1.83 & 0.8747$\pm$0.0467 & 29.56$\pm$7.14 \\
    & Neighbor2Neighbor & 37.27$\pm$1.76 & 0.8919$\pm$0.0444 & 27.99$\pm$6.47 \\
    & SDCNN & 38.03$\pm$1.22 & 0.9085$\pm$0.0299 & 25.34$\pm$3.90 \\
    & MM-Net & 39.01$\pm$1.41 & 0.9246$\pm$0.0309 & 22.71$\pm$4.09 \\
    & Noise2Sim & 39.44$\pm$1.44 & 0.9411$\pm$0.0229 & 21.53$\pm$3.81 \\
    & SVB & 39.69$\pm$1.23 & 0.9408$\pm$0.0247 & 21.02$\pm$3.24 \\
    & \textbf{Noise2Match (Ours)} & \textbf{41.02$\pm$1.56} & \textbf{0.9505$\pm$0.0234} & \textbf{18.08$\pm$3.58} \\
    \midrule
    \multirow{6}{*}{\rotatebox{90}{Piglet}} & LDCT & 36.58$\pm$2.03 & 0.9287$\pm$0.0193 & 30.08$\pm$6.55 \\
    &Noise2Void & 36.49$\pm$2.03 & 0.9280$\pm$0.0194 & 30.72$\pm$6.70 \\
    &Neighbor2Neighbor & 36.63$\pm$2.01 & 0.9335$\pm$0.0182 & 29.89$\pm$6.49 \\
    &SDCNN &  36.58$\pm$1.90 & 0.9452$\pm$0.0127 & 29.99$\pm$6.27 \\
    &MM-Net &  36.93$\pm$2.06 & 0.9475$\pm$0.0099 & 28.90$\pm$6.33 \\
    &Noise2Sim & 37.20$\pm$1.81 & 0.9643$\pm$0.0080 & 27.85$\pm$5.49 \\
    &SVB & 37.39$\pm$1.96 & 0.9665$\pm$0.0068 & 27.34$\pm$5.78 \\
    &\textbf{Noise2Match (Ours)} & \textbf{37.77$\pm$1.94} & \textbf{0.9702$\pm$0.0073} & \textbf{26.24$\pm$5.18} \\
    \bottomrule
\end{tabular}
}
\label{tab2}
\end{table}

\begin{figure}[!t]
\centerline{\includegraphics[width=\columnwidth]{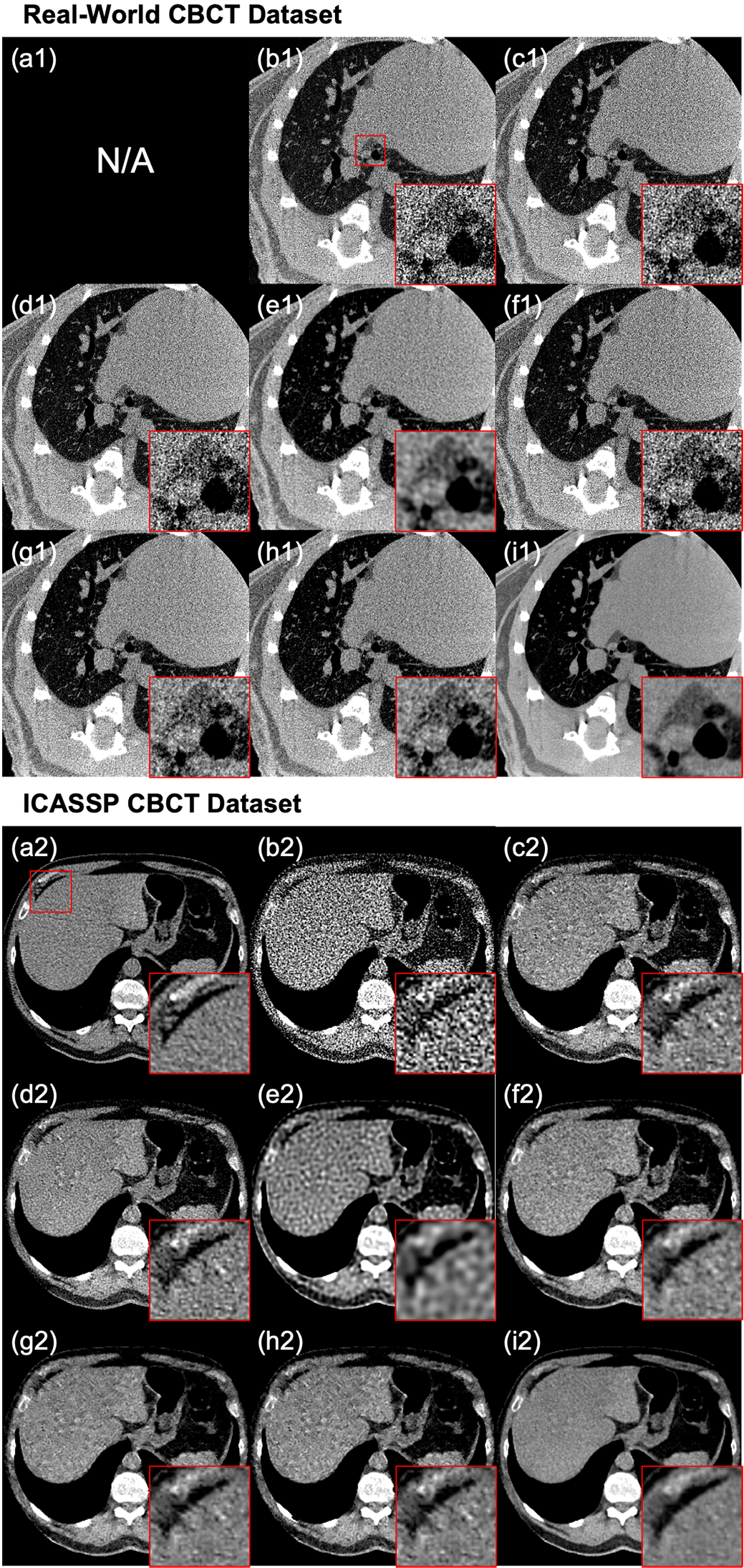}}
\caption{Qualitative results of different self-supervised methods on the Real-World CBCT and ICASSP CBCT datasets. (a) NDCT (ground truth; unavailable for the Real-World CBCT Dataset), (b) LDCT, (c) Noise2Void, (d) Neighbor2Neighbor, (e) SDCNN, (f) MM-Net, (g) Noise2Sim, (h) SVB, (i) Noise2Match (\textbf{ours}).}
\label{fig5}
\end{figure}

\begin{figure}[!t]
\centerline{\includegraphics[width=\columnwidth]{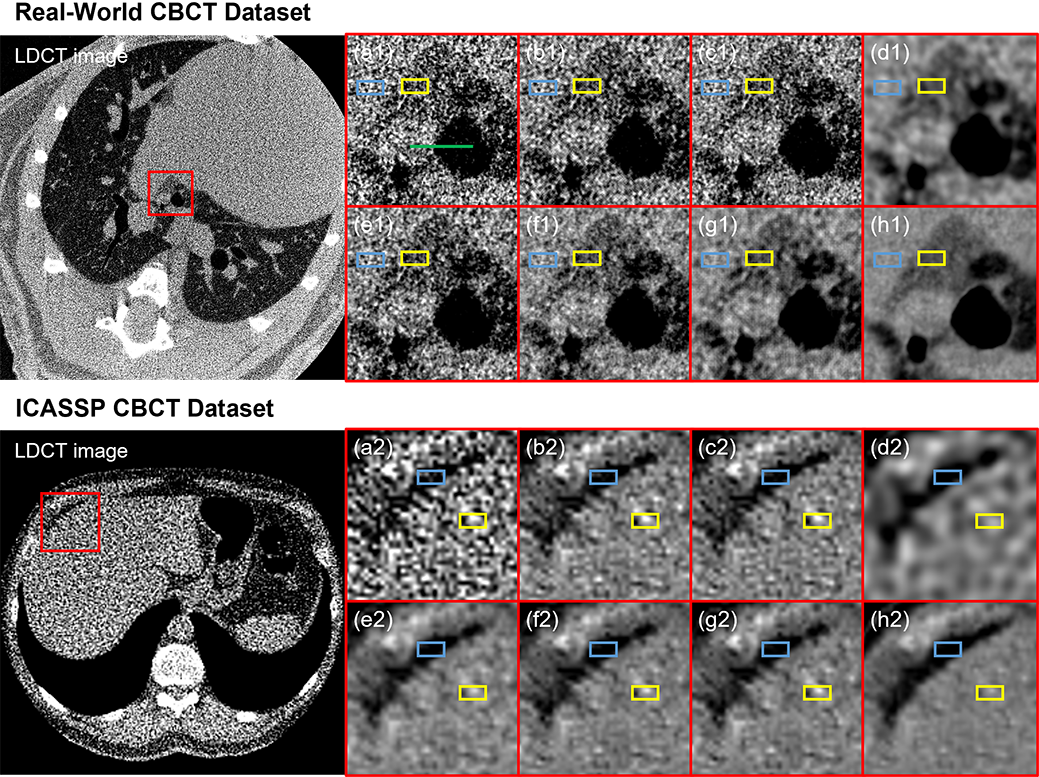}}
\caption{Qualitative results of selected ROIs for calculating CNR and Std (HU). The blue and yellow ROIs are used to calculate CNR, and the yellow ROIs are used to compute Std. (a) LDCT, (b) Noise2Void, (c) Neighbor2Neighbor, (d) SDCNN, (e) MM-Net, (f) Noise2Sim, (g) SVB, (h) Noise2Match (\textbf{ours}). }
\label{fig6}
\end{figure}

\begin{figure}[!t]
\centerline{\includegraphics[width=\columnwidth]{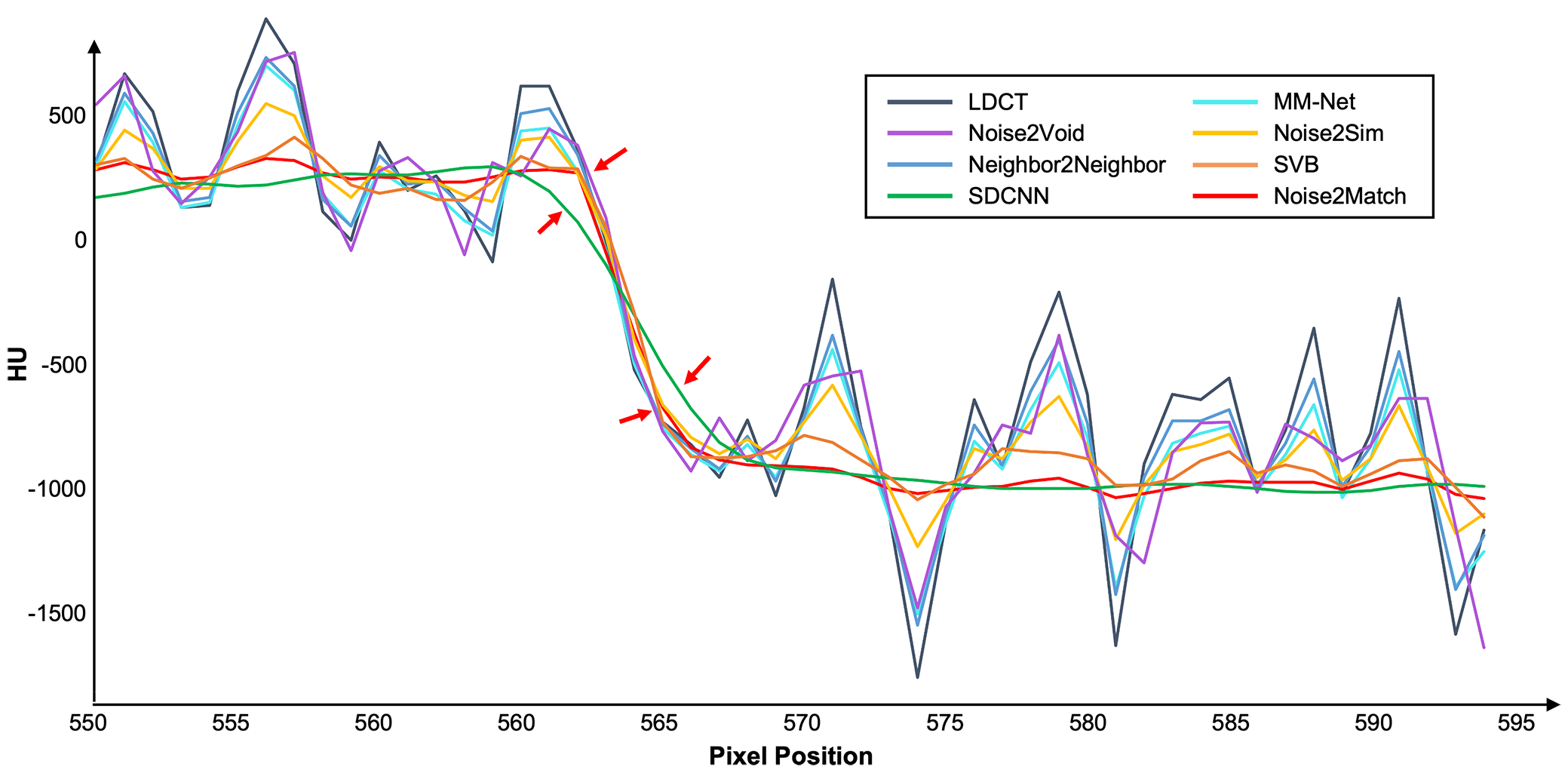}}
\caption{Line-profile comparison across the airway boundary indicated by the green line segment in Fig.~\ref{fig6}(a1).}
\label{fig7}
\end{figure}

\subsection{Performance Comparison on Low-Dose CBCT Datasets}
In this section, we evaluated the performance of various self-supervised methods on low-dose CBCT denoising. All self-supervised models were trained and tested on the Real-World CBCT and ICASSP CBCT datasets. Fig.~\ref{fig5} shows representative image results from different methods. It can be observed that Noise2Void and Neighbor2Neighbor failed to effectively suppress the noise in the CBCT images. SDCNN achieved only limited denoising performance on the CBCT datasets. On the Real-World CBCT dataset, noticeable residual noise remained in the denoised images, whereas on the ICASSP CBCT dataset, granular artifacts were observed in the denoised results. The SDCNN outputs also exhibited a certain tendency toward over-smoothing. The adjacent-slice-based baseline methods, including MM-Net, Noise2Sim, and SVB, also showed markedly degraded denoising performance on the CBCT datasets. This is likely because the noise in adjacent CBCT slices exhibits relatively strong correlation, which violates the assumption of inter-slice noise independence. Notably, although Theorem 1 also assumes independent noise between adjacent slices, Noise2Match was still able to effectively remove low-dose noise while preserving anatomical structures. This could be attributed to the correction operation during Noise2Match training, which can match pixels that were originally located at different spatial positions in adjacent slices and exhibited weaker noise correlation, thereby reducing the noise correlation between the corrected adjacent slices. In the subsequent Section~\ref{sec_correlation}, we further analyze the inter-slice noise correlation and verify the above deduction.

\begin{table*}[!t]
\centering
\caption{Quantitative results of different self-supervised methods on the Real-World CBCT and ICASSP CBCT datasets.}
\resizebox{0.8\textwidth}{!}{
\begin{tabular}{lccccccc}
     \toprule
     \multirow{2}{*}{Methods} & \multicolumn{2}{c}{Real-World CBCT} & \multicolumn{5}{c}{ICASSP CBCT} \\
     \cmidrule(r){2-3} \cmidrule(l){4-8}
     & CNR$\uparrow$ & Std (HU)$\downarrow$ & PSNR (dB)$\uparrow$ & SSIM$\uparrow$ & RMSE (HU)$\downarrow$ & CNR$\uparrow$ & Std (HU)$\downarrow$ \\
     \midrule
     LDCT & 0.73 & 94.63 & 29.30$\pm$0.66 & 0.7260$\pm$0.0378 & 67.45$\pm$6.60 & 0.90 & 138.06 \\
     Noise2Void & 0.92 & 76.30 & 31.33$\pm$0.97 & 0.8390$\pm$0.0243 & 54.59$\pm$5.63 & 1.46 & 82.92 \\
     Neighbor2Neighbor & 0.95 & 72.61 & 31.37$\pm$0.90 & 0.8466$\pm$0.0194 & 53.03$\pm$5.76 & 1.79 & 76.15 \\
     
     SDCNN & 2.66 & 32.64 & 31.29$\pm$1.16 & 0.8765$\pm$0.0255 & 53.97$\pm$7.87 & 3.87 & 41.33 \\
     
     MM-Net & 1.03 & 69.53 & 31.84$\pm$0.94 & 0.8655$\pm$0.0293 & 49.50$\pm$6.12 & 2.17 & 64.14 \\
     Noise2Sim & 1.45 & 47.82 & 32.35$\pm$0.89 & 0.9042$\pm$0.0224 & 47.49$\pm$5.24 & 2.42 & 58.39 \\
     SVB & 1.68 & 41.78 & 32.60$\pm$0.96 & 0.9062$\pm$0.0210 & 46.36$\pm$5.82 & 2.66 & 54.39 \\
     \textbf{Noise2Match (Ours)} & \textbf{4.53} & \textbf{13.29} & \textbf{34.14$\pm$1.21} & \textbf{0.9251$\pm$0.0172} & \textbf{38.90$\pm$5.80} & \textbf{5.23} & \textbf{26.95}\\
     \bottomrule
\end{tabular}
}
\label{tab3}
\end{table*}

Quantitative results are listed in Table~\ref{tab3}. Considering the lack of clean reference images in the Real-World CBCT dataset, we additionally introduced the Contrast-to-Noise Ratio (CNR) \cite{RA20} and Standard Deviation (Std) as supplementary quantitative measures of tissue contrast and noise suppression. Specifically, as shown in Fig.~\ref{fig6}, for the representative slices from the Real-World CBCT and ICASSP CBCT datasets presented in Fig.~\ref{fig5}, we further manually selected yellow and blue ROIs from two homogeneous regions. For each dataset, ROIs with identical sizes and spatial locations were applied to the LDCT image and the outputs of all competing methods. The ROIs were selected within homogeneous regions and away from visible tissue boundaries to reduce the influence of anatomical variations on the CNR and Std calculations. The yellow and blue ROIs were used to calculate CNR, while the yellow ROI was additionally used to calculate Std. Since CNR and Std are sensitive to ROI selection, the reported values were calculated only from the representative slices shown in Fig.~\ref{fig6} and are intended as supplementary quantitative measures to the qualitative comparison rather than as dataset-level performance metrics. Noise2Void and Neighbor2Neighbor achieved quantitative results only slightly better than those of the original LDCT images, whereas the adjacent-slice-based self-supervised baselines, including MM-Net, Noise2Sim, and SVB, generally achieved better performance. Although SDCNN obtained favorable CNR and Std values, its overall quantitative performance on the ICASSP CBCT dataset remained inferior to the stronger adjacent-slice-based baselines, particularly Noise2Sim and SVB, which may be related to its tendency toward over-smoothing. To further investigate this issue, we performed a line-profile analysis across an airway boundary along the green line segment indicated in Fig.~\ref{fig6}(a1), as shown in Fig.~\ref{fig7}. As indicated by the red arrows, the green curve corresponding to SDCNN exhibits a more gradual intensity transition across the airway boundary, providing further evidence of over-smoothing. In contrast, the red curve corresponding to Noise2Match maintains a relatively steep intensity transition, with a transition trend similar to that of the original LDCT image. This indicates that Noise2Match can effectively suppress noise while preserving the airway boundary without obvious over-smoothing. Overall, Noise2Match achieved the best quantitative performance on both CBCT datasets and showed clear advantages over the competing self-supervised methods. These quantitative and qualitative results demonstrate that Noise2Match can effectively perform low-dose CBCT denoising in a self-supervised manner while maintaining favorable tissue contrast, noise suppression, and structural preservation.

\subsection{Analysis of Inter-Slice Noise Correlation}
\label{sec_correlation}
\begin{figure}[!t]
\centerline{\includegraphics[width=\columnwidth]{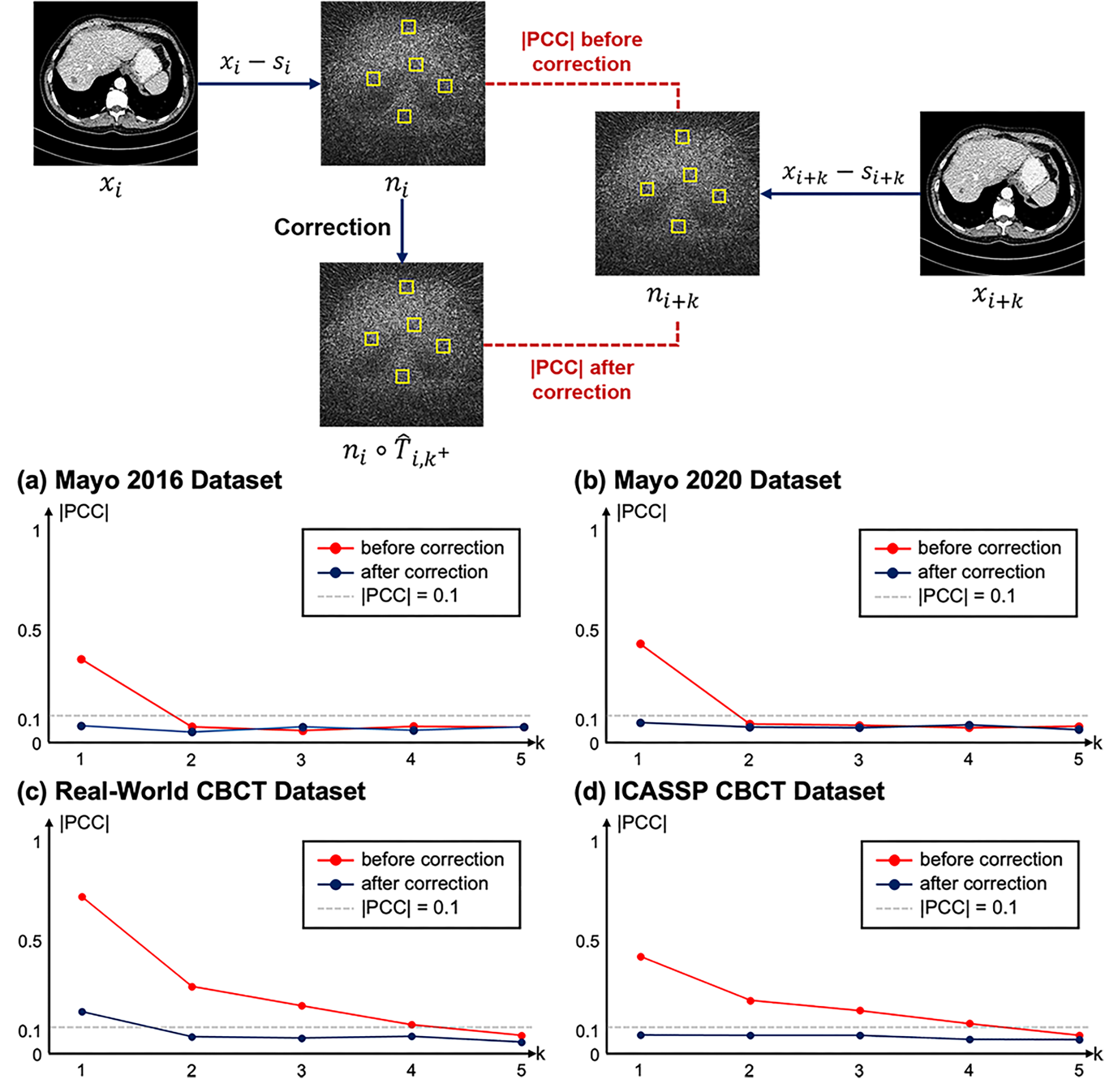}}
\caption{Quantitative analysis of the inter-slice noise correlation before and after the correction operation on the Mayo 2016, Mayo 2020, Real-World CBCT, and ICASSP CBCT datasets.}
\label{fig8}
\end{figure}

In this section, we quantitatively analyze the inter-slice noise correlation before and after the correction operation. This analysis was conducted on the Mayo 2016, Mayo 2020, Real-World CBCT, and ICASSP CBCT datasets. Utilizing the well-trained $f_\theta$ and $M_\phi$, we estimated the label transition matrices $\hat{T}_{i,k^+}=M_\phi(f_\theta(x_i),f_\theta(x_{i+k}))$. As illustrated in Fig.~\ref{fig8}, for every dataset and inter-slice distance $k\in \{1,2,3,4,5\}$, 100 slice pairs $(x_i,x_{i+k})$, each with five ROIs from different locations, were randomly selected from the training set for the calculation of the inter-slice noise correlation before and after correction. For the Mayo 2016, Mayo 2020, and ICASSP CBCT datasets, the noise map $n_i$ of the noisy slice $x_i$ was obtained as $n_i=x_i-s_i$, where $s_i$ denotes the corresponding clean reference image. For the Real-World CBCT dataset, since clean reference images are unavailable, we first applied Gaussian smoothing to the noisy CBCT image to estimate its low-frequency anatomical component, and then subtracted the smoothed image from the original noisy image to obtain an approximate noise map. ROIs in this dataset were selected only from homogeneous regions without visible anatomical edges to minimize the interference of residual anatomical structures in the noise correlation calculation.

For each selected ROI, the inter-slice noise correlation before correction was calculated as the absolute value of Pearson Correlation Coefficient ($|PCC|$) between the corresponding noise patches extracted from $n_i$ and $n_{i+k}$. After correction, the noise map $n_i$ was first warped according to the estimated transition matrix $\hat{T}_{i,k^+}$. The corrected inter-slice noise correlation was then calculated as the $|PCC|$ between the corresponding noise patches extracted from $n_i\circ\hat{T}_{i,k^+}$ and $n_{i+k}$. Each point in Fig.~\ref{fig8} represents the mean $|PCC|$ over 500 ROI-level measurements, corresponding to 100 selected slice pairs and 5 ROIs per pair. As shown by the red curves in Fig.~\ref{fig8}, when the inter-slice distance $k$ is small, both clinical fan-beam CT and CBCT exhibit non-negligible inter-slice noise correlation. However, the noise correlation in clinical fan-beam CT decays more rapidly as $k$ increases, whereas the noise correlation in CBCT persists across multiple neighboring slices, indicating stronger long-range inter-slice noise correlation. This observation may help explain why existing adjacent-slice-based self-supervised methods, especially those that attempt to reduce inter-slice noise correlation by increasing $k$, perform well on clinical CT data but are difficult to directly adapt to CBCT data. The blue curves in Fig.~\ref{fig8} show that, after the correction operation, the mean $|PCC|$ decreases substantially across all four datasets. These results demonstrate that the correction operation in Noise2Match training can effectively reduce inter-slice noise correlation, making the corrected slice pairs better aligned with the independent noise assumption required by Theorem 1.

To further examine whether the reduction in inter-slice noise correlation was caused by spatial resampling itself, we conducted an additional control experiment on the Real-World CBCT dataset with $k=2$, using the same ROI selection protocol as described above. The original inter-slice $|PCC|$ was 0.2935$\pm$0.0377 (mean$\pm$standard deviation). We then applied a fixed 0.5-pixel leftward translation to the noisy slice $x_i$ using bilinear interpolation. This operation introduced spatial resampling only, without correction based on inter-slice anatomical correspondence. The resulting $|PCC|$ was 0.2917$\pm$0.0414, which was nearly unchanged from the original value. In contrast, when $x_i$ was corrected using the estimated transition matrix $\hat{T}_{i,k^+}$, the $|PCC|$ decreased to 0.0752$\pm$0.0206 with bilinear interpolation. When the interpolation method was further replaced with nearest-neighbor interpolation, the $|PCC|$ similarly decreased to 0.0691$\pm$0.0181. These results indicate that spatial resampling or interpolation itself has little effect on inter-slice noise correlation, and that the substantial reduction in correlation after correction is mainly attributable to the meaningful inter-slice matching provided by $M_\phi$.

\subsection{Analysis of Learned Inter-Slice Deformation Fields}
\begin{figure}[!t]
\centerline{\includegraphics[width=\columnwidth]{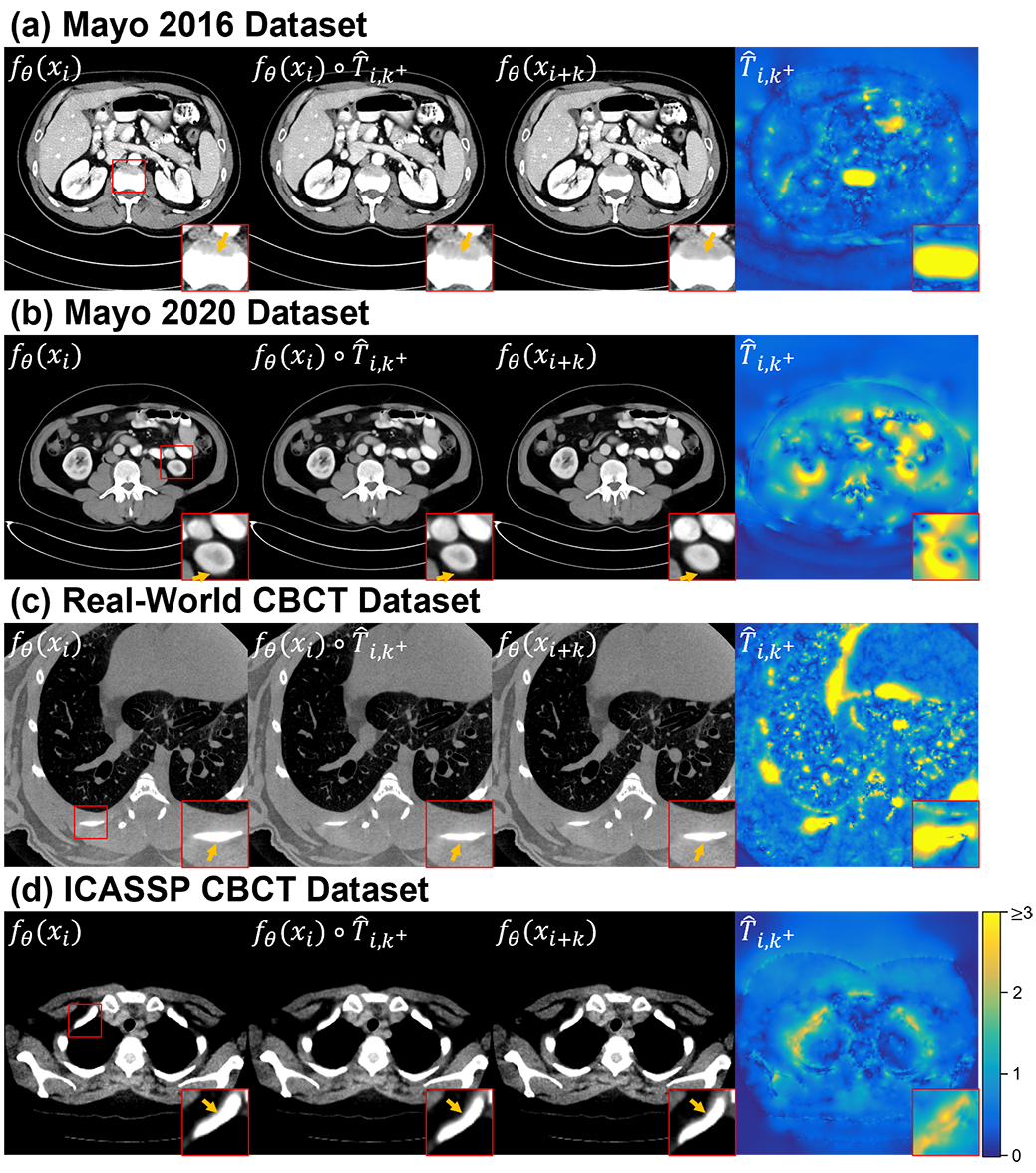}}
\caption{Representative visualization of the deformation fields learned by the inter-slice matching network on the Mayo 2016, Mayo 2020, Real-World CBCT, and ICASSP CBCT datasets.}
\label{fig9}
\end{figure}

To further investigate whether the inter-slice matching network $M_\phi$ can learn spatially regular and anatomically plausible transformations, we analyzed the deformation fields $\hat{T}_{i,k^+}$ estimated by $M_\phi$. Representative results on the Mayo 2016, Mayo 2020, Real-World CBCT, and ICASSP CBCT datasets are shown in Fig.~\ref{fig9}. For each pair of adjacent slices, we sequentially present the denoised current slice $f_\theta(x_i)$, the corrected current slice $f_\theta(x_i)\circ\hat{T}_{i,k^+}$, the denoised adjacent slice $f_\theta(x_{i+k})$, and the displacement magnitude map derived from $\hat{T}_{i,k^+}$, expressed in pixels. The displacement magnitude at spatial location $p$ is defined as

\begin{equation}
D_{i,k^+}(p)=\sqrt{(\hat{T}_{i,k^+}^{x}(p))^2+(\hat{T}_{i,k^+}^{y}(p))^2},
\label{eq18}
\end{equation}

\noindent where $\hat{T}_{i,k^+}^{x}$ and $\hat{T}_{i,k^+}^{y}$ denote the horizontal and vertical displacement components, respectively. As shown in Fig.~\ref{fig9}, the estimated deformation fields vary continuously across space, with relatively large displacements mainly distributed in regions exhibiting noticeable anatomical differences between adjacent slices. The enlarged regions further show that, after applying $\hat{T}_{i,k^+}$, the anatomical structures in $f_\theta(x_i)\circ\hat{T}_{i,k^+}$ exhibit better spatial correspondence with those in $f_\theta(x_{i+k})$, without introducing visually apparent structural distortions. These results indicate that $M_\phi$ mainly compensates for local spatial misalignments caused by anatomical variations between adjacent slices, and that the learned deformation fields $\hat{T}_{i,k^+}$ are anatomically plausible.

We further quantitatively evaluated the local spatial regularity of $\hat{T}_{i,k^+}$ using the Jacobian determinant. The spatial transformation corresponding to $\hat{T}_{i,k^+}$ is defined as

\begin{equation}
\phi(p)=p+\hat{T}_{i,k^+}(p),
\label{eq19}
\end{equation}

\noindent where $p$ denotes a spatial location. The Jacobian determinant $det(J_\phi)$ characterizes the local behavior of the spatial transformation, while $det(J_\phi)\leq0$ indicates local folding or orientation reversal. Accordingly, we define the folding ratio of $\hat{T}_{i,k^+}$ as the proportion of spatial locations with non-positive Jacobian determinants:

\begin{equation}
R_{fold}=\frac{|\{p\in\Omega:det(J_\phi(p))\leq0\}|}{|\Omega|}.
\label{eq20}
\end{equation}

\begin{table}
\centering
\caption{Jacobian statistics of the deformation fields estimated by the inter-slice matching network.}
\resizebox{0.8\columnwidth}{!}{
\begin{tabular}{lcc}
    \toprule
    Dataset & Jacobian determinant & Folding ratio \\
    \midrule
    Mayo 2016 & 0.9999$\pm$0.0495 & 0\%    \\
    Mayo 2020 & 1.0001$\pm$0.0727 & 0\%    \\
    Real-World CBCT & 1.0002$\pm$0.0604 & 0\%    \\
    ICASSP CBCT & 1.0007$\pm$0.0703 & 0\%    \\
    \bottomrule
\end{tabular}
}
\label{tab4}
\end{table}

\noindent Here, $\Omega$ denotes the set of all spatial locations in the deformation field. We calculated the Jacobian determinant at each spatial location for all estimated deformation fields in the four training sets, and computed the corresponding folding ratios. As shown in Table~\ref{tab4}, the mean and standard deviation of the Jacobian determinants were 0.9999$\pm$0.0495, 1.0001$\pm$0.0727, 1.0002$\pm$0.0604, and 1.0007$\pm$0.0703 on the Mayo 2016, Mayo 2020, Real-World CBCT, and ICASSP CBCT training sets, respectively. The Jacobian determinants on all four datasets were tightly concentrated around 1, indicating that the estimated transformations generally introduced only mild local expansion or contraction. More importantly, the folding ratio was 0\% on all four datasets, indicating that no local folding was observed in any of the evaluated deformation fields. These qualitative and quantitative results demonstrate that, under the constraints of the correction loss $\mathcal{L}_{Corr}$ and smoothness loss $\mathcal{L}_{Sm}$, the inter-slice deformation fields learned by $M_\phi$ are spatially regular and anatomically plausible.

\subsection{Ablation Study}
In this section, we conducted ablation studies to examine the effects of individual components in Noise2Match and different values of the parameters $k$ and $th$. The component ablation and the ablation on $th$ were conducted on the Mayo 2016 dataset, whereas the ablation on $k$ was conducted on the Mayo 2016, Mayo 2020, Real-World CBCT, and ICASSP CBCT datasets.

\subsubsection{Ablation on Different Components}
\begin{figure}[!t]
\centerline{\includegraphics[width=\columnwidth]{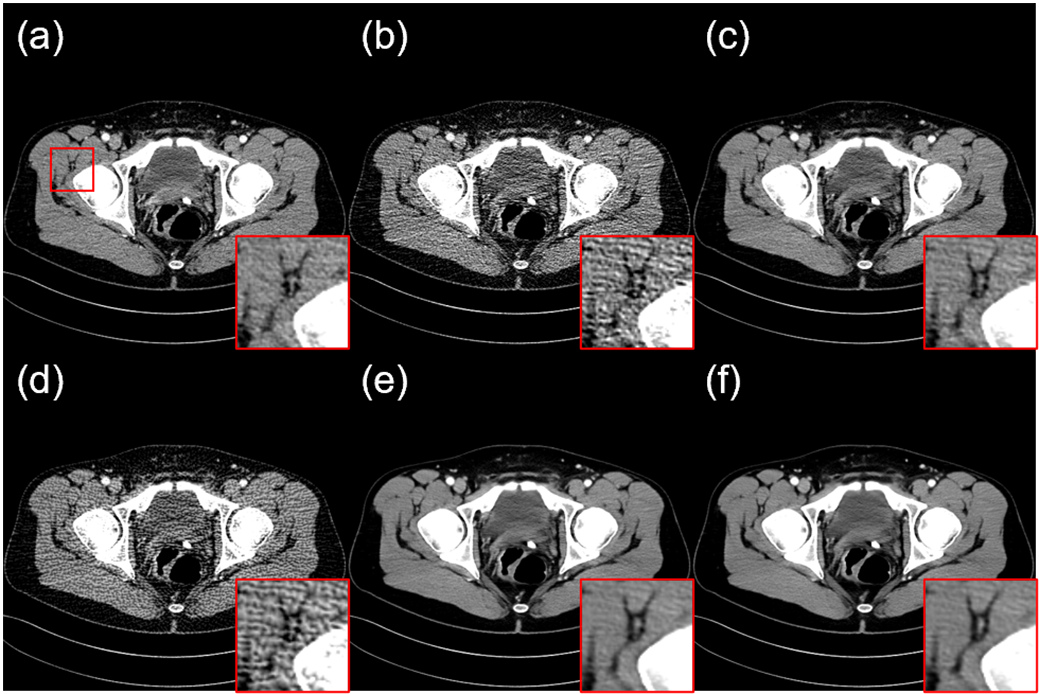}}
\caption{Qualitative ablation results of different components on the test data from Mayo 2016 Dataset. (a) NDCT, (b) LDCT, (c) w/o $M_\phi$, $\mathcal{L}_{Corr}$, $\mathcal{L}_{Sm}$, and $R.W.$, (d) w/o $\mathcal{L}_{Sm}$ and $R.W.$, (e) w/o $R.W.$, (f) Noise2Match.}
\label{fig10}
\end{figure}

We conducted ablation experiments on the Mayo 2016 Dataset to evaluate the effects of the proposed inter-slice matching network $M_\phi$, the correction loss $\mathcal{L}_{Corr}$, the smooth loss $\mathcal{L}_{Sm}$, and the loss reweighting mechanism ($R.W.$). Fig.~\ref{fig10} presents qualitative results, while Table~\ref{tab5} lists quantitative results. When all proposed components were removed, the denoiser was trained directly using the loss function in Eq.~\eqref{eq2}. As shown in Fig.~\ref{fig10}(c), residual noise remained in the denoised images, and the quantitative metrics dropped by 2.22 dB in PSNR, 0.0181 in SSIM, and increased by 4.94 HU in RMSE compared to the full Noise2Match listed in the fourth row of Table~\ref{tab5}. This indicates that directly training the model with adjacent slices leads to considerable interference in denoising performance due to mean differences caused by anatomical variation. When only $M_\phi$ and $\mathcal{L}_{Corr}$ were incorporated, without $\mathcal{L}_{Sm}$ and loss reweighting, the model performed even worse. Fig.~\ref{fig10}(d) shows noticeable granular noise, and the quantitative metrics in the second row of Table~\ref{tab5} were the lowest. We speculate that the absence of smoothness constraints allowed $M_\phi$ to output deformation fields with excessive distortions or folds that did not conform to the natural anatomical variations, thereby interfering with the optimization of the denoiser. Incorporating $M_\phi$, $\mathcal{L}_{Corr}$, and $\mathcal{L}_{Sm}$ together, with loss reweighting disabled (corresponding to $th\rightarrow \infty$), improved the denoising performance. Nevertheless, some fine structural details in the ROIs of Fig.~\ref{fig10}(e) remained slightly blurred compared to Fig.~\ref{fig10}(f). Finally, incorporating all the proposed components enabled the denoising model to achieve the highest-quality image results and optimal quantitative scores. These experiments demonstrate the effectiveness of the inter-slice matching network $M_\phi$, correction loss $\mathcal{L}_{Corr}$, smooth loss $\mathcal{L}_{Sm}$, and loss reweighting mechanism in our proposed Noise2Match.

\begin{table}
\centering
\caption{Quantitative ablation results of different components on the Mayo 2016 Dataset.}
\resizebox{\columnwidth}{!}{
\begin{tabular}{ccccccc}
     \toprule
     $M_\phi$ & $\mathcal{L}_{Corr}$ & $\mathcal{L}_{Sm}$ & $R.W.$ & PSNR (dB)$\uparrow$ & SSIM$\uparrow$ & RMSE (HU)$\downarrow$ \\
     \midrule
     - & - & - & - & 37.86$\pm$0.81 & 0.9229$\pm$0.0211 & 24.83$\pm$2.13 \\
     \checkmark & \checkmark & - & - & 35.90$\pm$0.91 & 0.8552$\pm$0.0299 & 32.22$\pm$3.53 \\
     \checkmark & \checkmark & \checkmark & - & 39.68$\pm$0.85 & 0.9368$\pm$0.0187 & 20.38$\pm$2.02 \\
     \checkmark & \checkmark & \checkmark & \checkmark & 40.08$\pm$0.84 & 0.9410$\pm$0.0167 & 19.89$\pm$1.96 \\
     \bottomrule
\end{tabular}
}
\label{tab5}
\end{table}

\subsubsection{Ablation on Different Parameter k}
\label{sec_k}
\begin{figure}[!t]
\centerline{\includegraphics[width=\columnwidth]{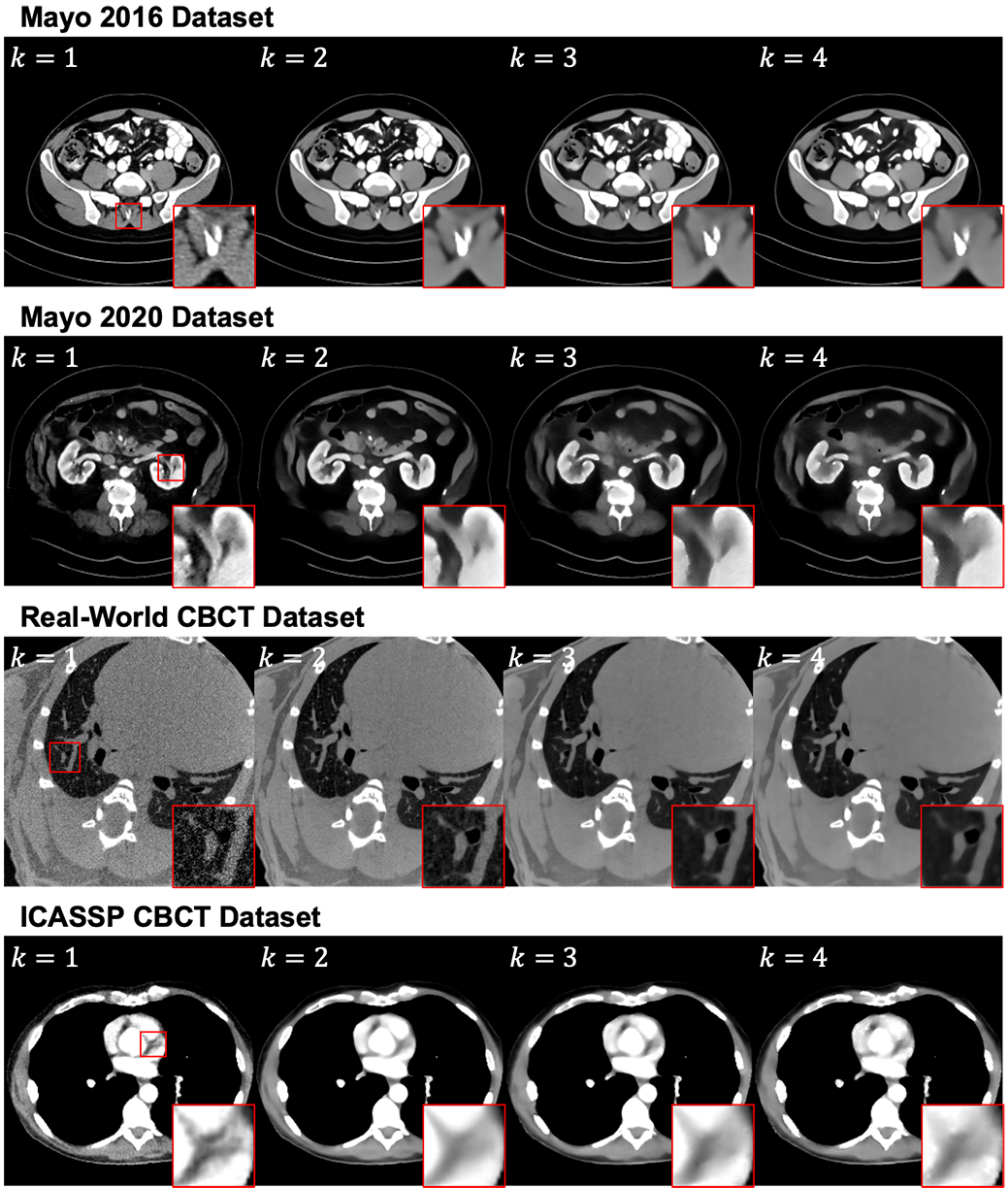}}
\caption{Qualitative ablation results of different parameter k on the Mayo 2016, Mayo 2020, Real-World CBCT and ICASSP CBCT datasets.}
\label{fig11}
\end{figure}

We investigated the effect of the inter-slice distance parameter $k$ on the performance of Noise2Match. Fig.~\ref{fig11} shows representative image results obtained with different $k$ values on the Mayo 2016, Mayo 2020, Real-World CBCT, and ICASSP CBCT datasets. Table~\ref{tab6} lists the quantitative results on the Mayo 2016, Mayo 2020, and ICASSP CBCT datasets. As shown in Fig.~\ref{fig11}, when $k=1$, Noise2Match achieves favorable image results on the Mayo 2016, Mayo 2020, and ICASSP CBCT datasets, effectively suppressing noise while preserving fine anatomical structures. However, on the Real-World CBCT dataset, noticeable residual noise can still be observed when $k=1$. This may be because the noise correlation between adjacent slices in this dataset remains relatively strong at $k=1$. The red curves in Fig.~\ref{fig8} also support this observation: when $k=1$, the inter-slice noise correlation before correction in the Real-World CBCT dataset is clearly higher than that in the other three datasets. When $k=2$, Noise2Match achieves the best visual performance on the Real-World CBCT dataset, with substantially reduced residual noise and well-preserved structural information. However, for the other three datasets, further increasing $k$ does not bring additional benefits, but instead leads to a certain degree of image smoothing. As $k$ further increases to 3 and 4, although noise can still be effectively suppressed, the image results on all datasets exhibit more pronounced over-smoothing, with many fine structural details being lost. The quantitative results in Table~\ref{tab6} are generally consistent with the visual observations in Fig.~\ref{fig11}. When $k=1$, the Mayo 2016, Mayo 2020, and ICASSP CBCT datasets all achieve the best quantitative scores. As $k$ increases, PSNR and SSIM gradually decrease on all three datasets, while RMSE gradually increases. These qualitative and quantitative results indicate that although a larger $k$ can help reduce inter-slice noise correlation, it also increases the structural differences between adjacent slices, making it more difficult for the inter-slice matching network to accurately match pixels belonging to the same tissues, thereby compromising the preservation of fine structural details.

\begin{table}[!t]
\centering
\caption{Quantitative ablation results of different parameter $k$ on the Mayo 2016, Mayo 2020 and ICASSP CBCT datasets.}
\resizebox{0.8\columnwidth}{!}{
\begin{tabular}{clccc}
    \toprule
     & $k$ & PSNR (dB)$\uparrow$ & SSIM$\uparrow$ & RMSE (HU)$\downarrow$  \\
    \midrule
    \multirow{4}{*}{\rotatebox{90}{\fontsize{7}{8}\selectfont Mayo 2016}} 
    & 1 & 40.08$\pm$0.84 & 0.9410$\pm$0.0167 & 19.89$\pm$1.96 \\
    & 2 & 39.37$\pm$0.96 & 0.9334$\pm$0.0172 & 21.92$\pm$2.27 \\
    & 3 & 38.59$\pm$0.97 & 0.9261$\pm$0.0156 & 23.64$\pm$2.64 \\
    & 4 & 38.10$\pm$1.18 & 0.9182$\pm$0.0187 & 25.10$\pm$3.52 \\
    \midrule
    \multirow{4}{*}{\rotatebox{90}{\fontsize{7}{8}\selectfont Mayo 2020}} 
    & 1 & 42.88$\pm$2.48 & 0.9648$\pm$0.0189 & 15.32$\pm$6.46 \\
    & 2 & 40.93$\pm$1.84 & 0.9587$\pm$0.0150 & 19.02$\pm$5.23 \\
    & 3 & 38.62$\pm$1.50 & 0.9428$\pm$0.0134 & 25.52$\pm$5.06 \\
    & 4 & 38.09$\pm$1.45 & 0.9411$\pm$0.0153 & 26.71$\pm$5.09 \\
    \midrule
    \multirow{4}{*}{\rotatebox{90}{\fontsize{7}{8}\selectfont ICASSP}} 
    & 1 & 34.14$\pm$1.21 & 0.9251$\pm$0.0172 & 38.90$\pm$5.80 \\
    & 2 & 33.93$\pm$1.12 & 0.9142$\pm$0.0215 & 39.76$\pm$5.39 \\
    & 3 & 33.31$\pm$0.98 & 0.9054$\pm$0.0230 & 42.59$\pm$5.05 \\
    & 4 & 33.06$\pm$0.90 & 0.8947$\pm$0.0252 & 43.84$\pm$4.99 \\
    \bottomrule
\end{tabular}
}
\label{tab6}
\end{table}

The above analysis shows that the selection of $k$ essentially involves a trade-off between inter-slice noise correlation and inter-slice structural differences. A larger $k$ generally reduces inter-slice noise correlation, making the constructed slice pairs better aligned with the noise-independence assumption. However, it may also introduce more pronounced anatomical differences between slices, making inter-slice matching more challenging. In practical applications, $k$ can be determined using the following two strategies. The first strategy is to measure the inter-slice noise correlation in homogeneous regions of the original images before model training, so that values of $k$ associated with excessively high inter-slice noise correlation can be excluded. This measurement does not require clean reference images. Specifically, the low-frequency anatomical components within homogeneous regions can first be approximated using Gaussian smoothing or a similar method and then subtracted from the original noisy images to obtain approximate noise maps, based on which the inter-slice noise correlation can be estimated. The second strategy is to adjust $k$ according to the visual quality of the model outputs on the training data. Specifically, if noticeable residual noise remains in the output images, it usually indicates that $k$ is too small and that the inter-slice noise correlation remains relatively high. In contrast, if obvious over-smoothing appears in the output images, it usually indicates that $k$ is too large and that the increased inter-slice structural differences between slices have adversely affected model training. Following these strategies, we set $k=1$ for the Mayo 2016, Mayo 2020, Piglet, and ICASSP CBCT datasets, and $k=2$ for the Real-World CBCT dataset.

\subsubsection{Ablation on Different Parameter th for Loss Reweighting}
The parameter $th$ is used to identify residual structural differences between adjacent slices after correction and to reduce their influence on model optimization. In practice, $th$ can be empirically selected using the training data by visualizing the resulting reweighting masks. Specifically, when the masks obtained over a range of $th$ values consistently highlight similar edge-like structures corresponding to residual inter-slice structural differences, we consider this range of $th$ values to be reasonable, and a specific $th$ value can then be empirically selected from this range. Based on this criterion, we set $th=80$ HU in our experiments.

\begin{figure}[!t]
\centerline{\includegraphics[width=\columnwidth]{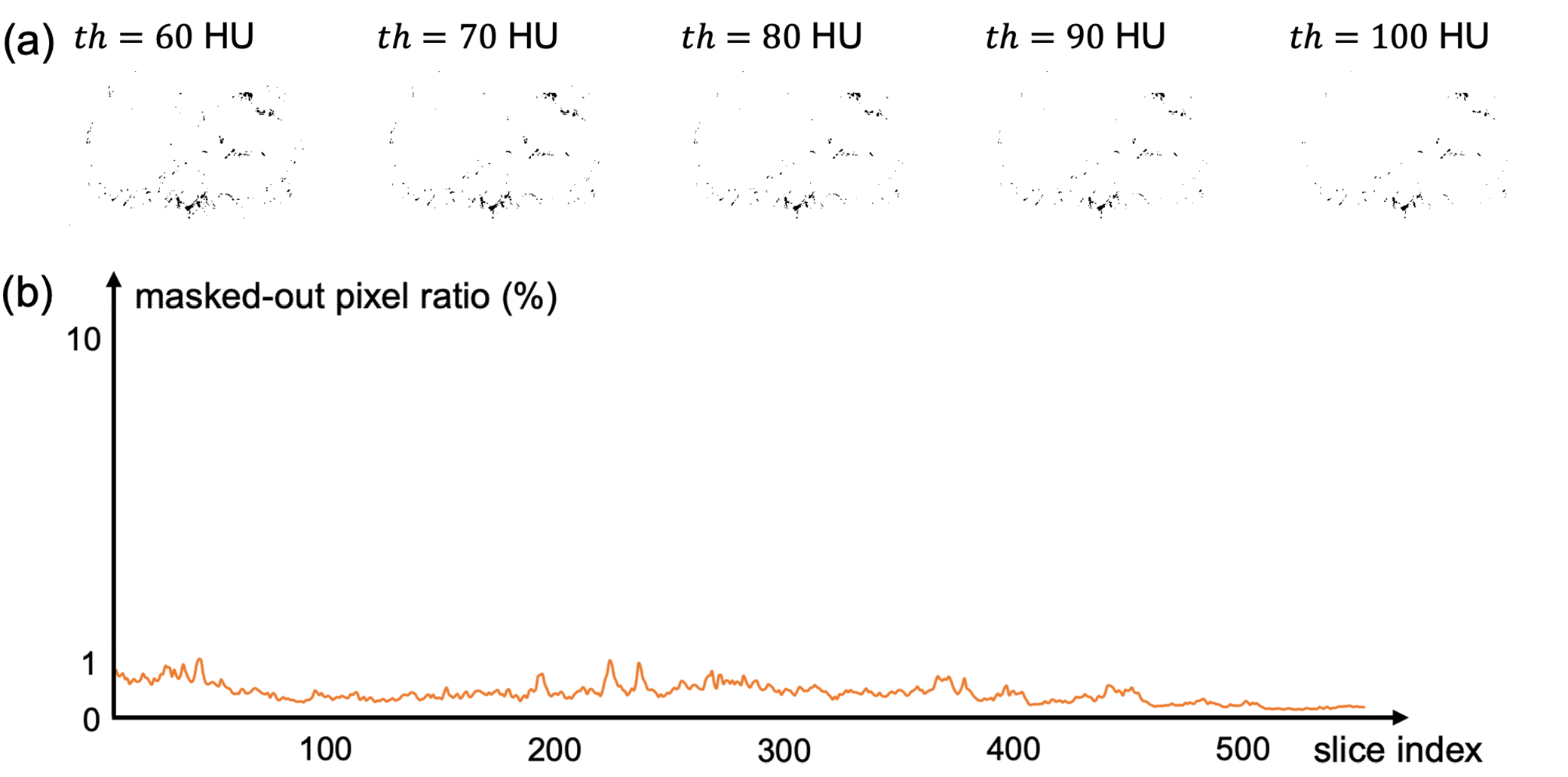}}
\caption{Analysis of the loss-reweighting mask on the Mayo 2016 training set. (a) Representative reweighting masks obtained with different $th$ values. (b) Slice-wise masked-out pixel ratio in a representative training volume using $th=80$ HU.}
\label{fig12}
\end{figure}

To evaluate the influence of different $th$ values for loss reweighting on the performance of Noise2Match, we conducted ablation experiments on the Mayo 2016 dataset. Fig.~\ref{fig12}(a) shows the corresponding reweighting masks for a representative pair of adjacent slices from the training set. It can be observed that, although the number of masked pixels varies slightly with $th$, the masks consistently highlight similar edge-like structures within this range, which is consistent with our criterion for selecting $th$. In addition, as shown in Table~\ref{tab7}, when $th$ varies from 60 HU to 100 HU, the quantitative performance of Noise2Match changes only slightly. These qualitative and quantitative results indicate that Noise2Match is not sensitive to the exact setting of $th$ within a reasonable range.

In addition, to quantify the proportion of pixels affected by loss reweighting during training, we selected a representative volume from the Mayo 2016 training set and, with $th=80$ HU, calculated for each slice the proportion of pixels excluded by the reweighting mask relative to the entire image. As shown in Fig.~\ref{fig12}(b), for most slices in this volume, the masked-out pixels account for only a very small proportion of the entire image, typically less than 1\%. This result indicates that loss reweighting mainly acts on a small number of regions with noticeable residual inter-slice structural differences.

\begin{table}[!t]
\centering
\caption{Quantitative ablation results of different parameter $th$ on the Mayo 2016 Dataset.}
\resizebox{0.8\columnwidth}{!}{
\begin{tabular}{lccc}
     \toprule
     $th$ & PSNR (dB)$\uparrow$ & SSIM$\uparrow$ & RMSE (HU)$\downarrow$ \\
     \midrule
     60 & 40.08$\pm$0.82 & 0.9406$\pm$0.0170 & 19.90$\pm$1.96 \\
     70 & 40.07$\pm$0.84 & 0.9409$\pm$0.0169 & 19.92$\pm$1.97 \\
     80 & 40.08$\pm$0.84 & 0.9410$\pm$0.0167 & 19.89$\pm$1.96 \\
     90 & 40.05$\pm$0.85 & 0.9411$\pm$0.0152 & 19.98$\pm$2.01 \\
     100 & 40.07$\pm$0.83 & 0.9407$\pm$0.0173 & 19.91$\pm$1.96 \\
     \bottomrule
\end{tabular}
}
\label{tab7}
\end{table}

\subsubsection{Empirical Evaluation of the Equivariance Condition}
\label{sec_equivariance}
\begin{figure}[!t]
\centerline{\includegraphics[width=\columnwidth]{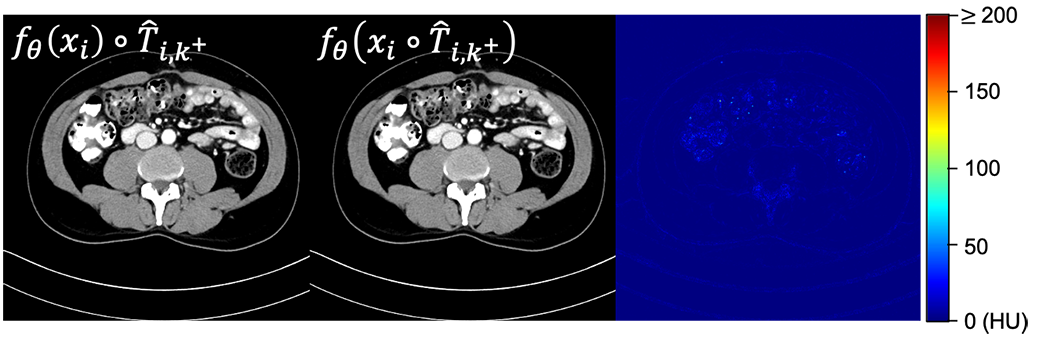}}
\caption{Empirical evaluation of the equivariance condition in Theorem 1 using a representative adjacent-slice pair from the Mayo 2016 training set. From left to right: $f_\theta (x_i)\circ\hat{T}_{i,k^+}$, $f_\theta (x_i \circ\hat{T}_{i,k^+})$, and the absolute difference between the two results.}
\label{fig13}
\end{figure}

The derivation of Theorem 1 relies on the equivariance condition of the denoising network with respect to the inter-slice deformation field, as this condition enables the corrected inter-slice training objective to be connected to the standard Noise2Noise objective. To examine how well this condition is satisfied in practical Noise2Match training, we performed a simple empirical evaluation. Specifically, we selected a representative pair of adjacent slices $(x_i,x_{i+k})$ from the Mayo 2016 training set and separately computed $f_\theta (x_i)\circ\hat{T}_{i,k^+}$ and $f_\theta (x_i \circ\hat{T}_{i,k^+})$, where $\hat{T}_{i,k^+}$ was estimated by $M_\phi$. Fig.~\ref{fig13} shows the image results obtained from the two computational paths. The two results are visually highly consistent, and only minor residual differences are observed in their difference map. These results indicate that, in practical Noise2Match training, $f_\theta$ approximately satisfies the equivariance condition required by Theorem 1 with respect to the estimated transition matrix $\hat{T}_{i,k^+}$.

\section{Discussion and Conclusion}

In this paper, we propose Noise2Match, a self-supervised LDCT denoising method based on adaptive inter-slice matching. Through theoretical analysis and experimental validation, we demonstrated that Noise2Match achieves superior denoising performance compared with existing state-of-the-art self-supervised methods, particularly for low-dose CBCT denoising. The reasons for Noise2Match’s outstanding performance can be summarized as follows. \textit{First}, in addition to using adjacent LDCT slices with high similarity and better noise independence as training data, Noise2Match employs a correction operation that allows the transfer of supervision signals between adjacent slices to adapt to non-rigid misalignments caused by anatomical variation. Consequently, compared with existing adjacent-slice-based self-supervised methods, Noise2Match has better satisfaction for the equal-mean condition required by the Noise2Noise framework. \textit{Second}, the correction operation in Noise2Match can match pixels that were originally at different positions in adjacent slices and had weaker noise correlation, further enhancing noise independence between slices. This enables Noise2Match to perform better low-dose CBCT denoising solely in the image domain, while existing self-supervised methods cannot accomplish this task well. \textit{Third}, Noise2Match incorporates a loss reweighting strategy, which mitigates the interference from mismatched pixels at longitudinal tissue boundaries between adjacent slices, further improving model performance.

It should be noted that the equivalence established in Theorem 1 relies on several idealized assumptions. In practical implementations, these assumptions can generally be satisfied only approximately because spatial transformations of discrete images involve interpolation and the transition matrix estimated by $M_\phi$ may contain errors. Therefore, in practice, the Noise2Match training objective should be understood as approximately, rather than strictly, equivalent to the supervised denoising objective. Although some theoretical assumptions can only be approximately satisfied in practice, the empirical evaluation in Section~\ref{sec_equivariance} and the experimental results across multiple datasets demonstrate that these approximations do not prevent Noise2Match from achieving effective and robust denoising performance.

In conclusion, in this work, we proposed Noise2Match, a self-supervised LDCT denoising method based on adaptive inter-slice matching. This method introduces an inter-slice matching network to model the label transition process between adjacent slices and applies correction to the denoiser’s outputs, enabling the adjacent slices to better satisfy the independent and equal-mean conditions. Additionally, a loss reweighting mechanism is employed to further enhance the model’s performance. Both theoretical analysis and extensive experiments on multiple datasets demonstrate that Noise2Match achieves superior performance over state-of-the-art image-domain self-supervised methods in various LDCT denoising tasks, particularly in low-dose CBCT denoising. Therefore, Noise2Match offers broader applicability and higher practical value.


\begin{thebibliography}{00}

\bibitem{SH19} H.~Shan \emph{et~al.}, ``{Competitive performance of a modularized deep neural network compared to commercial algorithms for low-dose CT image reconstruction},'' \emph{Nat. Mach. Intell.}, vol.~1, no.~6, pp. 269--276, 2019.

\bibitem{CH17} H.~Chen \emph{et~al.}, ``{Low-dose CT with a residual encoder-decoder convolutional neural network},'' \emph{IEEE Trans. Med. Imaging}, vol.~36, no.~12, pp. 2524--2535, 2017.

\bibitem{YQ18} Q.~Yang \emph{et~al.}, ``{Low-dose CT image denoising using a generative adversarial network with Wasserstein distance and perceptual loss},'' \emph{IEEE Trans. Med. Imaging}, vol.~37, no.~6, pp. 1348--1357, 2018.

\bibitem{LJ18} J.~Lehtinen \emph{et~al.}, ``{Noise2Noise: Learning image restoration without clean data},'' in \emph{{Proc. Int. Conf. Mach. Learn. (ICML)}}, 2018, pp. 2965--2974.

\bibitem{KA19} A.~Krull, T.-O. Buchholz, and F.~Jug, ``{Noise2Void: Learning denoising from single noisy images},'' in \emph{{Proc. IEEE/CVF Conf. Comput. Vis. Pattern Recognit. (CVPR)}}, 2019, pp. 2129--2137.

\bibitem{WZ22} Z.~Wang, J.~Liu, G.~Li, and H.~Han, ``{Blind2Unblind: Self-supervised image denoising with visible blind spots},'' in \emph{{Proc. IEEE/CVF Conf. Comput. Vis. Pattern Recognit. (CVPR)}}, 2022, pp. 2027--2036.

\bibitem{HT21} T.~Huang, S.~Li, X.~Jia, H.~Lu, and J.~Liu, ``{Neighbor2Neighbor: Self-supervised denoising from single noisy images},'' in \emph{{Proc. IEEE/CVF Conf. Comput. Vis. Pattern Recognit. (CVPR)}}, 2021, pp. 14781--14790.

\bibitem{LD25} D.~Liu, Y.~Shi, G.~Zhang, S.~Li, and L.~Zhang, ``{Zero-Shot Noise2Mean: Gap minimization for efficient denoising from a single noisy image},'' in \emph{{Proc. AAAI Conf. Artif. Intell.}}, vol.~39, no.~5, 2025, pp. 5406--5414.

\bibitem{NC23} C.~Niu \emph{et~al.}, ``{Noise suppression with similarity-based self-supervised deep learning},'' \emph{IEEE Trans. Med. Imaging}, vol.~42, no.~6, pp. 1590--1602, 2023.

\bibitem{ZY25} Y.~Zhu, Q.~He, Y.~Yao, and Y.~Teng, ``{Self-supervised Noise2Noise method utilizing corrupted images with a modular network for LDCT denoising},'' \emph{Pattern Recognit.}, vol. 161, p. 111285, 2025.

\bibitem{WQ23} Q.~Wu \emph{et~al.}, ``{Unsharp structure guided filtering for self-supervised low-dose CT imaging},'' \emph{IEEE Trans. Med. Imaging}, vol.~42, no.~11, pp. 3283--3294, 2023.

\bibitem{HY25} Y.~He, X.~Luo, C.~Wang, and W.~Yu, ``{Self-supervised learning for low-dose CT image denoising method based on guided image filtering},'' \emph{Phys. Med. Biol.}, vol.~70, no.~13, p. 135010, 2025.

\bibitem{SY25} Y.~Sun \emph{et~al.}, ``{Filter2Noise: A framework for interpretable and zero-shot low-dose CT image denoising},'' \emph{J. Med. Imag.}, vol.~13, no.~2, p. {024004}, 2026.

\bibitem{PJ24} J.~Pan, D.~Chang, W.~Wu, Y.~Chen, and S.~Wang, ``{Self-supervised tomographic image noise suppression via residual image prior network},'' \emph{Comput. Biol. Med.}, vol. 179, p. 108837, 2024.

\bibitem{ZH24} H.~Zhao, Y.~Gu, Z.~Zhao, B.~Du, Y.~Xu, and R.~Yu, ``{WIA-LD2ND: Wavelet-based image alignment for self-supervised low-dose CT denoising},'' in \emph{{Proc. Int. Conf. Med. Image Comput. Comput.-Assist. Intervent. (MICCAI)}}, 2024, pp. 764--774.

\bibitem{LY25} Y.~Liu \emph{et~al.}, ``{SDCNN: Self-supervised disentangled convolutional neural network for low-dose CT denoising},'' \emph{IEEE Trans. Instrum. Meas.}, vol.~74, pp. 1--13, {2025}.

\bibitem{JS23} S.-Y. Jeon, W.~Kim, and J.-H. Choi, ``{MM-Net: Multiframe and multimask-based unsupervised deep denoising for low-dose computed tomography},'' \emph{IEEE Trans. Radiat. Plasma Med. Sci.}, vol.~7, no.~3, pp. 296--306, 2023.

\bibitem{WY25} Y.~Wang \emph{et~al.}, ``{SVB: Self-supervised real CT denoising via similarity-based visual blind-spot scheme},'' \emph{IEEE Trans. Instrum. Meas.}, vol.~74, pp. 1--12, {2025}.

\bibitem{GE25} E.~Guo, Z.~Wang, Z.~Zhao, and L.~Zhou, ``{Imbalanced medical image segmentation with pixel-dependent noisy labels},'' \emph{IEEE Trans. Med. Imaging}, vol.~44, no.~5, pp. 2016--2027, 2025.

\bibitem{HK25} K.~Han \emph{et~al.}, ``{Region uncertainty estimation for medical image segmentation with noisy labels},'' \emph{IEEE Trans. Med. Imaging}, vol.~44, no.~12, pp. 5197--5207, 2025.

\bibitem{LZ25} Z.~Liao, S.~Hu, Y.~Xie, and Y.~Xia, ``{Instance-dependent label distribution estimation for learning with label noise},'' \emph{Int. J. Comput. Vis.}, vol. 133, no.~5, pp. 2568--2580, 2025.

\bibitem{WH21} H.~Wei, L.~Tao, R.~Xie, and B.~An, ``{Open-set label noise can improve robustness against inherent label noise},'' \emph{Adv. Neural Inf. Process. Syst.}, vol.~34, pp. 7978--7992, 2021.

\bibitem{ZL24} L.~Zhou and G.~Li, ``{Reliable multi-modal medical image-to-image translation independent of pixel-wise aligned data},'' \emph{Med. Phys.}, vol.~51, no.~11, pp. 8283--8301, 2024.

\bibitem{FDK84} L.~A. Feldkamp, L.~C. Davis, and J.~W. Kress, ``{Practical cone-beam algorithm},'' \emph{J. Opt. Soc. Am. A}, vol.~1, no.~6, pp. 612--619, 1984.

\bibitem{TD08} D.~J. Tward and J.~H. Siewerdsen, ``{Cascaded systems analysis of the 3D noise transfer characteristics of flat-panel cone-beam CT},'' \emph{Med. Phys.}, vol.~35, no.~12, pp. 5510--5529, 2008.

\bibitem{BJ11} J.~Baek and N.~J. Pelc, ``{Local and global 3D noise power spectrum in cone-beam CT system with FDK reconstruction},'' \emph{Med. Phys.}, vol.~38, no.~4, pp. 2122--2131, 2011.

\bibitem{ZH14} H.~Zhang, L.~Ouyang, J.~Ma, J.~Huang, W.~Chen, and J.~Wang, ``{Noise correlation in CBCT projection data and its application for noise reduction in low-dose CBCT},'' \emph{Med. Phys.}, vol.~41, no.~3, p. 031906, 2014.

\bibitem{CK23} K.~Choi, S.~H. Kim, and S.~Kim, ``{Self-supervised denoising of projection data for low-dose cone-beam CT},'' \emph{Med. Phys.}, vol.~50, no.~10, pp. 6319--6333, 2023.

\bibitem{LJ25} J.~Liu, G.~Li, Q.~Zhao, and S.~Luo, ``{Data regularization for streak artifacts removal in self-supervised micro-CT denoising},'' \emph{IEEE Trans. Radiat. Plasma Med. Sci.}, vol.~9, no.~5, pp. 627--638, 2025.

\bibitem{BG19} G.~Balakrishnan, A.~Zhao, M.~R. Sabuncu, J.~Guttag, and A.~V. Dalca, ``{VoxelMorph: A learning framework for deformable medical image registration},'' \emph{IEEE Trans. Med. Imaging}, vol.~38, no.~8, pp. 1788--1800, 2019.

\bibitem{MC17} C.~H. McCollough \emph{et~al.}, ``{Low-dose CT for the detection and classification of metastatic liver lesions: Results of the 2016 low dose CT grand challenge},'' \emph{Med. Phys.}, vol.~44, no.~10, pp. e339--e352, 2017.

\bibitem{Mayo2020} T.~R. Moen \emph{et~al.}, ``{Low-dose CT image and projection dataset},'' \emph{Med. Phys.}, vol.~48, no.~2, pp. 902--911, 2021.

\bibitem{YX18} X.~Yi and P.~Babyn, ``{Sharpness-aware low-dose CT denoising using conditional generative adversarial network},'' \emph{J. Digit. Imag.}, vol.~31, no.~5, pp. 655--669, 2018.

\bibitem{ICASSP} A.~Biguri and S.~Mukherjee, ``{Advancing the frontiers of deep learning for low-dose 3D cone-beam computed tomography (CT) reconstruction},'' in \emph{{Proc. IEEE Int. Conf. Acoust., Speech, Signal Process. Workshops (ICASSPW)}}, 2024, pp. 81--82.

\bibitem{LIDC-IDRI} K.~S. Mader, ``{The Lung Image Database Consortium image collection (LIDC-IDRI)},'' IEEE DataPort, 2021, doi: 10.21227/zce3-jp96.

\bibitem{RO15} O.~Ronneberger, P.~Fischer, and T.~Brox, ``{U-Net: Convolutional networks for biomedical image segmentation},'' in \emph{{Proc. Int. Conf. Med. Image Comput. Comput.-Assist. Intervent. (MICCAI)}}, 2015, pp. 234--241.

\bibitem{DK07} K.~Dabov, A.~Foi, V.~Katkovnik, and K.~Egiazarian, ``{Image denoising by sparse 3-D transform-domain collaborative filtering},'' \emph{IEEE Trans. Image Process.}, vol.~16, no.~8, pp. 2080--2095, 2007.

\bibitem{RA20} A.~Rodriguez-Molares \emph{et~al.}, ``{The generalized contrast-to-noise ratio: A formal definition for lesion detectability},'' \emph{IEEE Trans. Ultrason. Ferroelectr. Freq. Control}, vol.~67, no.~4, pp. 745--759, {2020}.

\end{thebibliography}
\end{document}